\documentclass[journal]{IEEEtran}

\ifCLASSOPTIONcompsoc
  \usepackage[nocompress]{cite}
\else
  \usepackage{cite}
\fi
\ifCLASSINFOpdf
\else
\fi

\pdfoutput=1

\usepackage{xcolor}
\usepackage{footnote}
\usepackage{multirow}
\usepackage{times}
\usepackage{epsfig}
\usepackage{graphicx}
\usepackage{amsmath}
\usepackage{mathtools}
\usepackage{amssymb}
\usepackage{fixltx2e}
\usepackage{color}
\usepackage{algpseudocode}
\usepackage[switch]{lineno} 

\usepackage{makecell}
\usepackage{footnote}
\usepackage{multicol}
\usepackage{subcaption}
\usepackage[linesnumbered,ruled,vlined]{algorithm2e}
\usepackage[]{threeparttable}

\usepackage{mathtools}
\usepackage{footmisc}
\usepackage{enumitem}
\usepackage{pifont}

\graphicspath{{./figures/}{./figures/draw/}}

\newcommand{\cmark}{\ding{51}}%
\newcommand{\xmark}{\ding{55}}%

\newcommand{\gcmark}{{\color{black}\ding{51}}}%
\newcommand{\rxmark}{{\color{black}\ding{55}}}%

\newcommand{\bh}{\mathbf{h}}

\newcommand{\bo}{\mathbf{o}}

\newcommand{\bu}{\mathbf{u}}

\newcommand{\bx}{\mathbf{x}}

\newcommand{\bz}{\mathbf{z}}

\newcommand{\bH}{\mathbf{H}}
\newcommand{\bI}{\mathbf{I}}

\newcommand{\bsh}{\boldsymbol{h}}

\newcommand{\bsz}{{\boldsymbol{z}}}

\newcommand{\bsmu}{\boldsymbol{\mu}}
\newcommand{\bsmui}{\boldsymbol{\mu}^i}
\newcommand{\bsmut}{\boldsymbol{\mu}^t}

\newcommand{\bhi}{\mathbf{h}^{{i}}}
\newcommand{\bht}{\mathbf{h}^{{t}}}
\newcommand{\bxi}{\mathbf{x}^{{i}}}
\newcommand{\bxt}{\mathbf{x}^{{t}}}

\newcommand{\bshi}{\boldsymbol{h}^{{i}}}
\newcommand{\bsht}{\boldsymbol{h}^{{t}}}
\newcommand{\bsxi}{\boldsymbol{x}^{{i}}}
\newcommand{\bsxt}{\boldsymbol{x}^{{t}}}

\newcommand{\bbE}{\mathbb{E}}

\newcommand{\bbR}{\mathbb{R}}

\newcommand{\oset}{\mathcal{O}}

\newcommand{\sign}{\mathrm{sign}}

\newcommand{\IJS}{I_{\text{JS}}}
\newcommand{\DJS}{D_{\text{JS}}\hspace{-1pt}}
\newcommand{\DSKL}{D_{\text{SKL}}\hspace{-1pt}}
\newcommand{\DKL}{D_{\text{KL}}\hspace{-1pt}}

\usepackage[pagebackref=True,breaklinks=true,letterpaper=true,bookmarks=false]{hyperref}

\begin{document}
%\linenumbers

\title{Multi-Modal Mutual Information Maximization:\\A Novel Approach for Unsupervised Deep Cross-Modal Hashing}

%\author{Michael~Shell,~\IEEEmembership{Member,~IEEE,}
%        John~Doe,~\IEEEmembership{Fellow,~OSA,}
%        and~Jane~Doe,~\IEEEmembership{Life~Fellow,~IEEE}% <-this % stops a space
%\IEEEcompsocitemizethanks{\IEEEcompsocthanksitem M. Shell was with the Department
%of Electrical and Computer Engineering, Georgia Institute of Technology, Atlanta,
%GA, 30332.\protect\\
%% note need leading \protect in front of \\ to get a newline within \thanks as
%% \\ is fragile and will error, could use \hfil\break instead.
%E-mail: see http://www.michaelshell.org/contact.html
%\IEEEcompsocthanksitem J. Doe and J. Doe are with Anonymous University.}% <-this % stops an unwanted space

%%%%%%%%%%%%%%%%%%%%%%%%%%%%%%%%%%%%%%%%%%%%
%%% uncomment for showing author list
\author{
Tuan Hoang, Thanh-Toan Do,  Tam V. Nguyen, Ngai-Man Cheung
\IEEEcompsocitemizethanks{
\IEEEcompsocthanksitem Tuan Hoang and Ngai-Man Cheung are with the Singapore University of Technology and Design (SUTD), Singapore.
%\protect\\
Email:\newline nguyenanhtuan\_hoang@mymail.sutd.edu.sg, ngaiman\_cheung@sutd.edu.sg

Thanh-Toan Do is with Monash University, Australia. ~Email: toan.do@monash.edu

Tam V. Nguyen is with the University of Dayton, United States. ~E-mail:tamnguyen@udayton.edu
% \IEEEcompsocthanksitem Tam V. Nguyen is with the University of Dayton, United States.
% E-mail: \protect\\{tamnguyen}@udayton.edu
}
}

%%%%%%%%%%%%%%%%%%%%%%%%%%%%%%%%%%%%%%%%%%%%%%%%%%

%{Shell \MakeLowercase{\textit{et al.}}: Bare Demo of IEEEtran.cls for Computer Society Journals}

\IEEEtitleabstractindextext{%

%\begin{IEEEkeywords}
%Computer Society, IEEE, IEEEtran, journal, \LaTeX, paper, template.
%\end{IEEEkeywords}
}

\maketitle

\IEEEdisplaynontitleabstractindextext

\IEEEpeerreviewmaketitle

% \begin{IEEEkeywords}
% Image search, Binary hashing, Spectral hashing, Cross-modality.
% \end{IEEEkeywords}

\begin{abstract}
In this paper, we adopt the maximizing mutual information (MI) approach to tackle the problem of unsupervised learning of binary hash codes for efficient cross-modal retrieval. We proposed a novel method, dubbed \textit{Cross-Modal Info-Max Hashing (CMIMH)}. First, to learn informative representations that can preserve both intra- and inter-modal similarities, we leverage the recent advances in estimating variational lower-bound of MI to maximizing the MI between the binary representations and input features and between binary representations of different modalities. By jointly maximizing these MIs under the assumption that the binary representations are modelled by multivariate Bernoulli distributions, we can learn binary representations, which can preserve both intra- and inter-modal similarities, effectively in a mini-batch manner with gradient descent. Furthermore, we find out that trying to minimize the modality gap by learning similar binary representations for the same instance from different modalities could result in less informative representations. Hence, balancing between reducing the modality gap and losing modality-private information is important for the cross-modal retrieval tasks. Quantitative evaluations on standard benchmark datasets demonstrate that the proposed  method consistently outperforms other state-of-the-art cross-modal retrieval methods.
\end{abstract}
% \vspace{-5pt}
\section{Introduction}
The last few years have witnessed an exponential surge in the amount of information available online in heterogeneous modalities, e.g., images, tags, text documents, videos, subtitles, etc. 
Thus, it is desirable to have a single efficient system that can facilitate large-scale multi-media searches. In general, this system should support both single and cross-modality searches, i.e., the system returns a set of semantically relevant results of all modalities given a query in any modality.
In addition, to be used in large scale applications, the system should have efficient storage and fast searching. Several cross-modality hashing approaches have been proposed to handle the above challenges, in both \textit{supervised} \cite{CRH,HTH,SMH,QCH,SePH,DCH,DLFH,SMFH,8425016,8331146,THN,DCMH,DDCMH,CMDVH,PRDH,DVSH,SRDMH,CMMN,CPAH,DSPOH,ACMR,WU2020107335} and \textit{unsupervised} \cite{CVH,PDH,IMH,CMFH,LSSH,ACQ,FSH,CRE,DMHOR,MSAE,CorrAE,UGACH,CALM,WANG2017249,9165953} manners. Furthermore, as the unsupervised hashing does not require any label information, it is suitable for large-scale retrieval problem in which the label information is mostly unavailable. Thus, in this work, we focus on the unsupervised setting of the cross-modality hashing problem for retrieval tasks.

When learning binary representations for the cross-modal retrieval task, it is essential to preserve both intra- and inter-modal similarities in a common Hamming space. 
Equivalently, the binary representations should satisfy several requirements:
\textit{(i)} First, the representations necessarily capture information from the input features, i.e., preserve intra-modal similarity. 
% (ii) Second, for the image representations to effectively retrieve text samples, the image representations should capture as much information about the text modality as possible. Analogously, the text representation should contain as much information about the image modality as possible to retrieve image samples effectively. 
% \textit{(ii)} Second, for the representations of a modality to effectively retrieve samples of other modalities, the representations should capture as much information about other modalities as possible, i.e., preserve inter-modal similarity.
% \textit{(iii)} Third, to minimize the modality gap (i.e., heterogeneous gap) so that the similarity of between different modalities can be measured directly, the representations of different modalities should be well aligned with each other, i.e., binary codes from different modalities of the same instance should be as similar as possible \cite{DJSRH,UDCMH,FSH}.
% \textit{(iii)} Third, to minimize the modality gap (i.e., heterogeneous gap); the representations of different modalities should be well aligned with each other, i.e., binary codes from different modalities of the same instance should be as similar as possible \cite{DJSRH,UDCMH,FSH}. Minimizing the modality gap is necessary for the similarity between different modalities to be measured directly.
%
{\textit{(ii)} For the representations of a modality to effectively retrieve samples of other modalities (i.e., the inter-modal similarity is preserved), the representations of this modality should capture as much information about other modalities as possible.}  
Additionally, \textit{(iii)} the modality gap (i.e., heterogeneous gap) between the representations of different modalities should be minimized, i.e., binary codes of all modalities should be in the same common space and binary codes from different modalities of the same instance (which contain the same information) should be as similar as possible \cite{DJSRH,UDCMH,FSH}. Minimizing the modality gap is necessary for the similarity between different modalities to be measured directly. 

To preserve both intra- and inter-modal similarities in the unsupervised setting, many existing cross-modality hashing methods, both CNN-based and non CNN-based, relied on similarity matrices/graphs (one for each modality \cite{CVH,IMH,LCMH,ACQ,UDCMH} or to a joint similarity matrix for all modalities \cite{FSH}). Then they learn hash codes via the eigenvalue decomposition of the similarity matrices. 
However, constructing the similarity matrix could be challenging and computationally expensive for large datasets. Furthermore, eigenvalue decomposition decreases the mapping quality substantially when increasing the hash code length \cite{RFDH}.
Matrix Factorizaton (MF) based methods could avoid the large scale graph constructing and eigen-decomposition process by finding a shared latent semantic space \cite{CMFH,RFDH} that can reconstruct input data well for all modalities. 
%However, MF could be sensitive to outliners. Moreover, only simple and capability-limited linear projection is used in MF-based methods.
{
However, only simple and capability-limited linear projection is used in MF-based methods. In addition, scaling up MF-based methods for much larger datasets is non-trivial.
Recently, Li et. al. proposed Unsupervised coupled Cycle generative
adversarial Hashing networks (UCH) \cite{UCH}, which used pair-coupled generative adversarial networks (GAN) to learn representations for individual modality and generate compact hash codes. Even though, this approach can achieve very competitive performance, training the minimax loss of GAN can be challenging. 
}

% Taking a different approach, inspired by recent advances in unsupervised representation learning by maximizing an estimate of the mutual information (MI) \cite{infomax,MIB}; in this paper, we propose to learn informative binary representations for unsupervised cross-modal hashing via maximizing mutual information (MI). 
% We learn to preserve the intra-modal and inter-modal similarities via maximizing the MI between representations and input and the MI between representations of different modalities. More specifically, by assuming the binary representations to be modeled by multivariate Bernoulli distributions, we can maximizing the MI effectively using gradient descent optimization in mini-batch manner via maximizing their variational lower-bounds \cite{IM_algo,InfoGAN,f-gan,MINE,Nguyen_2010}. 

Taking a different approach, inspired by recent advances in unsupervised representation learning \cite{infomax,CPC,AMDIM,MIB}; in this paper, we propose to learn informative binary representations for unsupervised cross-modal hashing via maximizing mutual information (MI). 
We learn to preserve the intra-modal and inter-modal similarities via maximizing the MI between representations and input and the MI between representations of different modalities. 
{
More specifically, by adopting the Variational Information Maximization method \cite{IM_algo}, we can use the binary representations to be modeled by multivariate Bernoulli distributions. As a result, the binary representations can be learned easily and effectively by maximizing the MI between themselves and input via maximizing the estimated variational MI lower-bounds \cite{IM_algo,InfoGAN,f-gan,MINE,Nguyen_2010} using gradient descent optimization in the mini-batch manner. 
}

Furthermore, we find out that trying to minimize modality gap by learning similar binary representations for the same instance from different modalities could result in undesirable side-effect. Specifically, the modality-private information (i.e., the information of one modality that does not share with any other modality) is discarded. Consequently, the representations may become less representative for the input. Hence, balancing between reducing modality gap and losing modality-private information is important for the cross-modal retrieval tasks. 
% In addition, 

In addition to the above requirements for cross-modal retrieval tasks, independence and balance are well-known to be important properties of informative hash codes \cite{spectral_hashing,ITQ,SSH,SAH,8801918,HOANG2020102852}. The independence property,  i.e.,  different  bits  in the binary codes are independent to each other, is to ensure hash codes do not capture redundant information. The balance property, i.e., each bit has a $50\%$ chance of being {$0$ or $1$}, is to ensure hash codes contain a maximum amount of information \cite{SSH}.
By assuming the binary representations to be modeled by multivariate Bernoulli distributions, we propose to leverage the Total Correlation (TC) \cite{TC} as a regularizer (i.e., minimizing TC) to enhance the independence between hash bits. Furthermore, the balanced property can also be achieved by regularizing the Bernoulli distributions such that the averaged probabilities over a training set for a bit to be {$0$ or $1$} are equal and equal to $50\%$.  

{
In summary, by adopting the maximizing mutual information approach, we propose a novel framework, dubbed \textit{Cross-Modal Info-Max Hashing (CMIMH)}, whose main contributions are:
\begin{itemize}%[leftmargin=*]
    \item We propose to adopt the maximizing MI approach to learn binary representations for the cross-modal retrieval tasks. Besides maximizing MI between the representations and inputs, we explicitly maximize the MI between representations of different modalities, which is important to learn informative representations for cross-modal retrieval tasks. 
    % Furthermore, by adopting Bernoulli distributions, we can effectively learn informative binary representations using MI maximization.
    \item We find out that minimizing modality gap by learning similar binary representations for the same instance from different modalities could result in less informative representations. Since both informative representations and modality gap are important for cross-modal retrieval tasks, properly balancing these two factors is important to achieve good performance, as shown in our experiments. { To the best of our knowledge, our work is the first work that provide in-depth analyses about the trade-off between these two factors.}
    \item We propose to leverage the Total Correlation (TC) as a regularizer to enhance the independence between hash bits. 
    The experimental results confirm that minimizing TC results in more independence hash bits and higher performance.
    \item We compare our proposed method against various state-of-the-art unsupervised cross-modality hashing methods on three standard cross-modal benchmark datasets, i.e., MIR-Flickr25K, NUS-WIDE, and MS-COCO. Quantitative results justify our contributions and demonstrate that CMIMH outperforms the compared methods on various evaluation metrics and settings.
\end{itemize}
}
%\vspace{-4pt}
\section{Related works}
\label{sec:related_works}
%Firstly, we briefly discuss noticeable methods proposed for \textbf{unsupervised cross-modal hashing}, to which our method belongs.
In this section we briefly discuss noticeable methods proposed cross-modal hashing. 

\textbf{Supervised Cross-modal hashing.}
Supervised hashing methods can explore the semantic information to enhance the data correlation from different modalities (i.e., reduce modality gap) and reduce the semantic gap. Many supervised cross-modal hashing methods with shallow architectures have been proposed, for instance Co-Regularized Hashing (CRH) \cite{CRH}, Heterogeneous Translated Hashing (HTH) \cite{HTH}, Supervised Multi-Modal Hashing (SMH) \cite{SMH}, Quantized Correlation Hashing (QCH) \cite{QCH}, Semantics-Preserving Hashing (SePH) \cite{SePH}, Discrete Cross-modal Hashing (DCH) \cite{DCH}, and Supervised Matrix Factorization Hashing (SMFH) \cite{SMFH}. 
% The label information in SePH \cite{SePH} after being encoded into an affinity matrix is transformed to a probability distribution from which approximate hash codes are learned while minimizing the Kullback-Leibler divergence.
% In Discrete Cross-modal Hashing (DCH) \cite{DCH}, the authors iteratively approximate the optimal binary codes (i.e., learning one bit at a time)
% %bit by bit in a close form 
% to achieve the discriminative unified and the modality-specific binary codes under the classification framework.
All of these methods are based on hand-crafted features, which cannot effectively capture heterogeneous correlation between different modalities and may therefore result in unsatisfactory performance. Unsurprisingly, recent deep learning-based works \cite{THN,DCMH,DDCMH,CMDVH,SSAH,PRDH,DVSH,8331146} can capture heterogeneous cross-modal correlations more effectively.
Deep cross-modal hashing (DCMH) \cite{DCMH} simultaneously conducts feature learning and hash code learning in a unified framework. 
Pairwise relationship-guided deep hashing (PRDH) \cite{PRDH}, in addition, takes intra-modal and inter-modal constraints into consideration. 
Deep visual-semantic hashing (DVSH) \cite{DVSH} uses CNNs, long short-term memory (LSTM), and a deep visual semantic fusion network (unifying CNN and LSTM) for learning isomorphic hash codes in a joint embedding space. 
However, the text modality in DVSH is only limited to sequence texts (e.g., sentences). 
In Cross-Modal Deep Variational Hashing \cite{CMDVH,Liong2020DeepVA}, the authors first proposed to learn shared binary codes from a fusion network, then learn generative modality-specific networks for encoding out-of-sample inputs.
In Cross-modal Hamming Hashing \cite{CMHH}, the author proposed Exponential Focal Loss which puts higher losses on pairs of similar samples with Hamming distance much larger than 2 (in comparison with the sigmoid function with the inner product of binary codes). 
Mandal \textit{et al.} \cite{8425016} proposed Generalized Semantic Preserving Hashing (GSPH) which can work for unpaired inputs (i.e., given a sample in one modality, there is no paired sample in other modality.).
Song  \textit{et al.} \cite{CMMN} took advantage of the memory mechanism to design a memory network that can learn to store supporting information and retrieve the necessary information in reference. 
Xie \textit{et al.} \cite{CPAH} proposed  Multi-Task Consistency-Preserving Adversarial Hashing (CPAH), which consists of two modules: consistency refined module to learn modality-common and modality-private representations and multi-task adversarial learning module to  preserve the semantic consistency information between different modalities. 
Ji \textit{et. al.} \cite{CMZSH}  proposed a attribute-guided network (AgNet) framework to narrow the semantic gap brought by modality heterogeneity and category migration for the zero-shot cross-modal retrieval.

Although supervised hashing typically achieves very high performance, it requires a labor-intensive process to obtain large-scale labels, especially for multi-modalities, in many real-world applications.
% Therefore, our work focuses on the unsupervised cross-modality hashing.
% In the following, we review related unsupervised cross-modal hashing methods.
% Although supervised hashing methods achieve higher accuracy than unsupervised methods, they require the label information to supervise the training. 
In contrast, unsupervised hashing does not require any label information. Hence, it is suitable for large-scale image search in which the label information is usually unavailable.

% \textbf{Unsupervised Cross-modal hashing:}
% Unlike the supervised cross-modal retrieval, the unsupervised cross-modal retrieval receives less attention due to the difficulty of fusing various modalities .
\textbf{Unsupervised Cross-modal hashing.}
Cross-view hashing (CVH) \cite{CVH} and Inter-Media Hashing (IMH) \cite{IMH} adopt Spectral Hashing \cite{spectral_hashing} for the cross-modality hashing problem. These two methods, however, produce different sets of binary codes for different modalities, which may result in limited performance.
Linear cross-modal hashing (LCMH) \cite{LCMH} reduces the training complexity of IMH by representing training data with some cluster centers to avoid the large-scale graph construction process. 
In Predictable Dual-View Hashing (PDH) \cite{PDH}, the authors introduced the {predictability} to explain the idea of learning linear hyper-planes that each one divides a particular space into two subspaces represented by $-1$ or $1$. The hyper-planes, in addition, are learned in a self-taught manner, i.e., to learn a certain hash bit of a sample by looking at the corresponding bit of its nearest neighbors. 
Collective Matrix Factorization Hashing (CMFH) \cite{CMFH} aims to find consistent hash codes from different views by collective matrix factorization. 
% Also by using matrix factorization, 
Latent Semantic Sparse Hashing (LSSH) \cite{LSSH} was proposed to learn hash codes in two steps: first, latent features from images and texts are jointly learned with sparse coding, and then hash codes are achieved by using matrix factorization.
Subsequently, Wang \textit{et al.} \cite{RFDH} proposed Robust and Flexible Discrete Hashing (RFDH) which directly optimizes and generates the unified binary codes for various views in the unsupervised manner via matrix factorization such that large quantization errors caused by relaxation can be relieved to some extent. 
Inspired by CCA-ITQ \cite{ITQ}, Go \textit{et al.} \cite{ACQ} proposed Alternating Co-Quantization (ACQ) to alternately minimize the binary quantization error for each of modalities. 
In \cite{FSH}, the authors applied Nearest Neighbor Similarity \cite{NSS} to construct Fusion Anchor Graph (FSH) from text and image modals for learning binary codes. 
Recently, proposed Collective Affinity Learning Method (CALM) \cite{CALM}, which collectively and adaptively learns hashing functions in an unsupervised manner with an anchor graph constructed on partial multi-modal data.
We would like to refer readers to \cite{cross_retrieval_survey} for a more comprehensive survey on non-DNN-based cross-modal retrieval methods.

In addition to the aforementioned shallow methods, several works \cite{DMHOR,MSAE,CorrAE} utilized (stacked) auto-encoders for learning binary codes. These methods try to minimize the distance between hidden spaces of modalities to the preserve inter-modal semantic to a certain extent. Deep Binary Reconstruction (DBRC) \cite{DBRC} proposed to minimize the reconstruction error based on the shared binary representation. DBRC additionally proposed a scalable $\tanh$ activation with a learnable parameter, which can mitigate the gradient problem of the discrete domain of $\{-1, 1\}$ during training.
Recently, Zhang \textit{et al.} \cite{MGAH} propsoed Multi-pathway Generative Adversarial Hashing (MGAH), which consists of a generative model and a discriminative model. In which, the generative model fits the distribution over the manifold structure and selects informative data of other modalities. While the discriminative model learns to discriminate data generated from the generative model and data sampled a correlation graph (a graph which captures the underlying manifold structure across different modalities).
% first designed a graph-based unsupervised correlation method to capture the underlying manifold structure across different modalities, and a generative adversarial network to learn the manifold structure. 
% first designed a correlation graph to capture the underlying manifold structure across different modalities
Wu \textit{et al.} \cite{UDCMH} proposed  Unsupervised Deep Cross Modal Hashing (UDCMH), %which alternatively between feature learning, binary latent representation learning and hash function learning.  
that enables the feature learning to be jointly optimized with the binarization. 
Su \textit{et al.} \cite{DJSRH} proposed a joint-semantics affinity matrix, which integrates the neighborhood information of two modals, for mini-batch samples to train deep network in an end-to-end manner. 
% This method is named as Deep Joint Semantics Reconstructing Hashing (DJSRH).
% Additionally, inspired by Anchor Graph \cite{AnchorGraph}, we also propose a novel method to achieve more informative similarity graphs for the multi-modal context. Moreover, both UDCMH \cite{UDCMH} and DJSRH \cite{DJSRH} are only proposed for two modalities, and their abilities to adapt to the general case of $M$ modalities are uncertain, while our method is proposed for $M$ modalities.
% In Cycle-Consistent Deep Generative Hashing (CYC-DGH) \cite{CYC-DGH}, the authors leverage the cycle-consistent loss \cite{CycleGAN2017} to learn hashing functions that map inputs of different modalities into a common hashing space without requiring paired inputs. 
Li et. al. proposed Unsupervised coupled Cycle generative
adversarial Hashing networks (UCH) \cite{UCH}, which used pair-coupled generative adversarial networks to learn representations for individual modality and generate compact hash codes. 
Given a multi-modal \textit{unpaired} data, Wu \textit{et al.} \cite{CYC-DGH} adopted the cycle-consistent loss \cite{CycleGAN2017} to learn hashing functions. 
Although these methods make great progresses, the performance of these systems still has room for improvement.

\smallskip
\textbf{Representation learning with Mutual Information:} In MIHash \cite{MIHash}, the authors proposed to use mutual information (MI) to learn hash codes for online hashing. In specific, given two Hamming distance distributions of a sample with its neighbors and its non-neighbors, the MI is used to measure the separability of these two distributions, which gives a good quality indicator for online hashing. However, this method is only proposed for the single modality case in a supervised manner, while our proposed method aims to learn binary representations for cross-modal retrieval in an unsupervised manner.
In \cite{disentangle_rep_with_MI}, the authors only adopted MI to learn to preserve intra-modal similarity, while our proposed method utilized MI to preserve both intra-modal and inter-modal similarities. Additionally, in contrast with \cite{disentangle_rep_with_MI} which learns real-valued representations, our method learns binary representations for large-scale retrieval.
Besides, several works \cite{infomax,ICP,AMDIM,MIB,CRD,CMC} rely on MI to learn representations in unsupervised/self-supervised manner for single view and/or multi-view settings.
Specifically, Hjelm \textit{et. al.} \cite{infomax} proposed to learn a global image representation such that the MI between this global representation and local features are maximized. Bachman \textit{et. al.} \cite{AMDIM} further improved \cite{infomax} by maximizing the MI between a global representation and local features of the different views (i.e., different images generated by data augmentation). 
Different from \cite{infomax,AMDIM}, \cite{VIB} adopted the Information Bottleneck (IB) objective \cite{IB_theory} with a variational approximation to minimize the MI between the input and the representation, while still ensuring the global representations can fulfill the target task (e.g., classification). This learning method could result in more robust representations. In \cite{MIB}, with the assumption that any single view can fully contain information of labels of a down-stream task (e.g., classification), the authors aimed to learn robust representations by capturing the shared information between views and discarding the private information (i.e., information that is exclusively contained in a particular view).
Information Competing Process (ICP) \cite{ICP} is another intriguing MI-based representation learning method. ICP aims to learn diversified representations by first separating a representation into two parts with different MI constraints, and then forcing separated parts to accomplish the downstream task independently without any knowledge of what the other part has learned.
However, these works \cite{infomax,ICP,AMDIM,MIB,CRD,CMC} mainly focus on learning a single real-valued representation from single/multi-view inputs for classification. In contrast, our work aims to learn binary representations for multi-modal retrieval.
% \vspace{-10pt}
%\clearpage

\section{Proposed method}
\label{sec:proposed_method}

Given a multi-modality dataset of $N$ instances, denoted as $\oset=\{\bo_j\}_{j=1}^N$, in which each instance is described by an image-text pair $\bo_j=(\bxi_j, \bxt_j)$, where $\bxi_j\in \bbR^{D^i}$ and $\bxt_j\in \bbR^{D^t}$ are the $j$-th  $D^i, D^t$ dimensional features of image and text modalities respectively. We aim to learn the corresponding $L$-bit binary representations $\bhi_j$ and $\bht_j\in\{0,1\}^L$ for each image and text pair $(\bxi_j, \bxt_j)$.
% We also denote the feature matrix for image and text modalities as $\bxi = \{\bXi_j\}_{j=1}^N \in \bbR^{N\times D^i}$ and $\bXt = \{\bxt_j\}_{j=1}^N \in \bbR^{N\times D^t}$. 푗
% We now delve into the details of our proposed method.

For representations being suitable for the cross-modal retrieval task, the representations should satisfy several requirements: 
(i) First, the representations should well represent the input data, i.e., they necessarily capture information from the input features. 
(ii) Second, for the image representations to effectively retrieve text samples, the image representations should capture as much information about the text modality as possible. Analogously, the text representation should contain as much information about the image modality as possible to retrieve image samples effectively. 
(iii) Third, the representations of different modalities should be well aligned with each other (i.e., the modal gap is minimized).

%\subsection{Maximize Mutual Information}
\subsection{Mutual Information Maximization}

Mutual information (MI) has been proven to be an important quantity in data science to measure the dependence of two random variables, since it can capture non-linear statistical dependencies between variables \cite{MINE}. Recent representation learning methods \cite{AMDIM,infomax} showed that MI maximization between inputs and encoder outputs can help to learn informative representations.
Hence to achieve the first requirement, we aim to maximize the MI between the binary representations and the input data. 
Noticeably, in the ideal case, when the representations fully capture all input information; the MI between image and text representations would be maximized and be equal to the MI between the image and text input data (which is a constant). Equivalently, the second requirement would be satisfied. However, in practice, the representations may not fully capture all input information. Hence, we propose to further enforce the second requirement by explicitly maximizing the MI between the representations of image and text modalities.
Our initial objective now can be written as follows:
% \vspace{-2pt}
\begin{equation}
\label{eq:primiary_objective}
    \max I(\bxi;\bhi) + I(\bxt;\bht) + I(\bhi;\bht).
%\vspace{-2pt}
\end{equation}
However, MI is well-known to be notoriously difficult to compute.
% particularly in continuous and high-dimensional settings. 
To handle this trouble, we propose to assume the image and text representations to be random variables; so that we can leverage recent advances in estimating variational lower bounds of MI \cite{IM_algo,InfoGAN,f-gan,MINE,Nguyen_2010} to maximize the objective function \eqref{eq:primiary_objective}. 

%\vspace{-2pt}
\subsection{Variational Lower Bounds of MI}
\subsubsection{Variational Information Maximization} Directly optimizing $I(\bxi;\bhi)$ and $I(\bxt;\bht)$ in the objective \eqref{eq:primiary_objective} is infeasible as the true {posterior distributions} (i.e., $P(\bxi|\bhi), P(\bxt|\bht)$) requiring for computing the MI is still unknown.
Fortunately, we can use the Variational Information Maximization \cite{IM_algo,InfoGAN} to compute the MI lower bound, in which $Q_{\phi_i}(\bxi|\bhi)$ can be used to approximate the true posterior distribution, as follows:
%\vspace{-3pt}
\begin{equation}
\label{eq:mi_img}
\begin{split}
    % I(\bxi,\bhi)&=H(\bxi)-H(\bxi|\bhi)\\
    %                       &=H(\bxi)+\bbE_{x^{(i)}\sim p(\bxi)}\left[\bbE_{\hat{h}^{(i)}\sim P(h^{(i)}|x^{(i)})} [\log{P(x^{(i)}|\hat{h}^{(i)})}]\right]
    I(\bxi;\bhi)&=H(\bxi)-H(\bxi|\bhi)\\
     &=H(\bxi)+\bbE_{p(\bxi)}\big[\bbE_{P_{\theta_i}(\bhi|\bxi)} [\log{P(\bxi|\bhi)}]\big]\\                        
     &=H(\bxi)+\bbE_{p(\bxi)}\big[\DKL[P(\bxi|\bhi)||Q_{\phi_i}(\bxi|\bhi)]\\
     &\qquad\qquad\qquad\qquad\quad+\bbE_{P_{\theta_i}(\bhi|\bxi)} [\log{Q_{\phi_i}(\bxi|\bhi)}]\big]\\
     &\ge H(\bxi)+\bbE_{p(\bxi)}\big[\bbE_{P_{\theta_i}(\bhi|\bxi)} [\log{Q_{\phi_i}(\bxi|\bhi)}]\big],
\end{split}
\end{equation}
where $H(\cdot)$ is the entropy function of a random variable; $\bbE$ is expectation, and $\theta_i;\phi_i$ represent the model parameters of the encoder and decoder distributions, respectively. Similarly, we have
\begin{equation}
\label{eq:mi_txt}
I(\bxt;\bht) \ge H(\bxt)+\bbE_{p(\bxt)}\big[\bbE_{P_{\theta_t}(\bht|\bxt)} [\log{Q_{\phi_t}(\bxt|\bht)}]\big].
\end{equation}
Note that $H(\bxi)$ and $H(\bxt)$ are constant for the given input data. To be concise, from now on, we skip the subscript about model parameters in the encoder and decoder distributions whenever the context is clear.

As we aim to obtain binary representations for the cross-modal hashing, we adopt the multivariate Bernoulli distributions to model the encoder distributions $P(\bhi|\bxi)$ and $P(\bht|\bxt)$; i.e., $P(\bhi|\bxi):=\text{Bern}(\bsmui)$ and $P(\bht|\bxt):=\text{Bern}(\bsmut)$.
% $P(\bhi|\bxi)\sim \text{Bern}(\bsmu) = \prod_{l=1}^L \mu_l^{\bhi_l}(1-\mu_l)^{(1-\bhi_l)}$ where $\bsmu = [\mu_l]_{l=1}^L \in [0, 1]^L$
Additionally, we assume that the decoder distributions $Q(\bxi|\bhi)$ and $Q(\bxt|\bht)$ are Gaussian. Therefore, the log likelihoods in \eqref{eq:mi_img} and \eqref{eq:mi_txt} can be maximized by minimizing $L2$ reconstruction loss. 

\subsubsection{Reparameterization trick}
Following \cite{concrete,REBAR}, we can reparameterize $\bh\sim \text{Bern}(\bsmu)$ as $\bh = \sign(\bz)$ (i.e., $\sign(z) = 1$ if $z \ge 0$ and $\sign(z) = 0$ if $z < 0$), where $\bsz$ is a vector of independent logistic random variables defined as follows
%\vspace{-1pt}
\begin{equation}
    \bz = g(\bu,\bsmu) = \log\frac{\bsmu}{1-\bsmu}+\log\frac{\bu}{1-\bu},
%\vspace{-1pt}
\end{equation}
where $\bu\sim \text{Uniform}(0,1)$.
% (i.e., $\sign(z_i) = 1$ if $z_i \ge 0$ and $\sign(z_i) = 0$ if $z_i < 0$, where $z_i$ is the $i$-th element of $\bsz$). 
Even though this reparameterization trick can help to avoid sampling from the Bernoulli distribution, 
% As a result,  However, being different from the reparameterization trick for  the Gaussian distribution, 
this trick still requires a discrete threshold function which hinders the gradient descent optimization. 
To handle this difficulty, we resort to the Straight-Through Estimator (STE) \cite{STE}, i.e., $\frac{\partial\sign(z)}{\partial z}=1$,
% _{|z|\le 1} i.e., $\frac{\partial \mathcal{L}}{\partial \bh}=\frac{\partial \mathcal{L}}{\partial \bz}$ for $\bS^w\in[-1, 1]$,
to approximate the gradients propagating through the $\sign$ function.
% To avoid the discrete problem, one could rely on the unbiased gradient estimators based on REINFORCE trick \cite{reinforce}. However, the REINFORCE trick suffers from very high variances. Although some variance reduction remedies have been proposed \cite{VIMCO,muprop,REBAR}, they require extra computation in practice. More importantly, as discussed in \cite{muprop}, the simple biased  can     

\subsubsection{Sample-based differentiable MI lower bound}
\label{sec:IJS}
Different from $I(\bxi;\bhi)$ and $I(\bxt;\bht)$; in $I(\bhi;\bht)$, we can access samples from two random variables independently. This allows us to maximize the MI between the two representations $I(\bhi;\bht)$ using a sample-based differentiable MI lower bound, which could have a tighter bound  than \eqref{eq:mi_img} and \eqref{eq:mi_txt} in practice \cite{VBMI}.
Furthermore, as our primary interest is to maximize the MI, and not to find its precise value; we can rely on non-KL divergences estimator, i.e., a Jensen-Shannon MI estimator ($\IJS$), which is observed to work better in practice (e.g., more stable) than the KL divergences MI estimator (e.g., Donsker-Varadhan representation (DV) \cite{Donsker-Varadhan} or $f$-divergences \cite{f-gan}) \cite{infomax}. The sample-based differentiable Jensen-Shannon MI estimator ($\IJS$) could be defined as follows
%
% \begin{equation}    
% \label{eq:mi_js}
% \begin{split}
% \hspace{-4pt}I(\bhi;\bht)\hspace{-1pt}\propto
% \IJS&(\bhi;\bht)\hspace{-1pt}=\hspace{-1pt}\DJS\left[p(\bhi, \bht)\big\|p(\bhi) p(\bht)\right] \\
%     &\ge\hspace{-1pt}\sup_{T\in\mathcal{F}} \bbE_{p(\bhi, \bht)}[t] \hspace{-1pt}+\hspace{-1pt}\bbE_{p(\bhi) p(\bht)}[\log(2\hspace{-1pt}-\hspace{-1pt} \exp(t))],\hspace{-4pt}
% \end{split}
% \end{equation}
%\vspace{-1pt}
\begin{align}
\label{eq:mi_js}
\hspace{-4pt}I(\bhi;\bht)\hspace{-1pt}\propto
\IJS&(\bhi;\bht)=\DJS\left[p(\bhi, \bht)\big\|p(\bhi) p(\bht)\right] 
\end{align}  
\vspace{-0.8cm}
\begin{align}
 &\ge\hspace{-1pt}\sup_{T\in\mathcal{F}} \bbE_{p(\bhi, \bht)}[\bar{T}] \hspace{-1pt}+\hspace{-1pt}\bbE_{p(\bhi) p(\bht)}[\log(2- \exp(\bar{T}))], \nonumber   
%\vspace{-1pt}
\end{align}    
where $\mathcal{F}$ is a family of functions $T(\bshi;\bsht) \in \bbR$, parameterized by neuron networks, which is jointly optimized during the training procedure to classify if a pair of samples are from the joint distribution $p(\bhi, \bht)$ 
% \footnote{$p(\bhi,\bht) = p(\bxi, \bxt)P(\bhi|\bxi)P(\bht|\bxt)$} 
or the product of marginal distributions $p(\bhi)p(\bht)$, i.e., pairs of $(\bshi;\bsht)$ are produced from the pairs of inputs sampled from joint distributions $p(\bxi,\bxt)$ or sampled from the product of marginal distributions $p(\bxi)p(\bxt)$, respectively
% \footnote{$p(\bhi) = p(\bxi)P(\bhi|\bxi)$ and $p(\bht) = p(\bxt)P(\bht|\bxt)$}
\cite{f-gan,MINE}. Additionally, $\bar{T}=\log(2)-\log(1+\exp(-T))$ \cite{f-gan}.

From \eqref{eq:mi_js}, we can see that if the function $T$ can correctly classify between samples from the joint and product of marginal distributions with high confident, the MI lower bound will be maximized. However, we found that the process of jointly training the function $T$ (with binary inputs sampled from $\bsh^i\sim\text{Bern}(\bsmui)$ and $\bsh^t\sim\text{Bern}(\bsmut)$) and the encoders may result in an undesirable side-effect.
Particularly, besides encouraging the encoders to learn the hidden variables for different modalities, such that the MI of these hidden variables is maximized; the function $T$ also promotes the encoders to reduce the stochasticity in $\text{Bern}(\bsmui)$ and $\text{Bern}(\bsmut)$ (i.e., $\bsmu\to 0$ or $\bsmu\to 1$). 
% by producing the Bernoulli variables to be very small (i.e., $\bsmu\to 0$) or large (i.e., $\bsmu\to 1$). 
Intuitively, reducing stochasticity in $\text{Bern}(\bsmui)$ and $\text{Bern}(\bsmut)$ (i.e., less noise) allows the function $T$ to correctly classify samples easier. 
Consequently, this side-effect may impact on the variational information maximization in \eqref{eq:mi_img} and \eqref{eq:mi_txt} as the $P(\bhi|\bxi)$ and $P(\bht|\bxt)$ become more deterministic\footnote{Note that the MI lower bound in \eqref{eq:mi_img} is derived for random variables \cite{IM_algo,InfoGAN}, and may not be applicable for deterministic variables.}.
Arguably, using multiple pairs of $(\bshi,\bsht)$
% of $\text{Bern}(\bsmu^{i})$ and $\text{Bern}(\bsmu^{t})$
could help to mitigate this problem. However, this requires a higher computational cost. 
% To effectively address the problem, we propose to directly use the Bernoulli variables $\bsmui$ and $\bsmut$ as the inputs for the function $T$, i.e., $V(\bsmui;\bsmut)$ instead of $T(\bhi;\bht)$. 
To effectively address the problem, we propose to directly classify if pairs of multivariate Bernoulli distributions $(\text{Bern}(\bsmui), \text{Bern}(\bsmut))$ (from which a pair $(\bshi,\bsht)$ is sampled) are produced from the pairs of inputs sampled from joint distributions $p(\bxi,\bxt)$ or sampled from the product of marginal distributions $p(\bxi)p(\bxt)$, i.e., $T(\bsmui;\bsmut)$ instead of $T(\bshi;\bsht)$.
This would help to eliminate noise in the inputs of the function $T$, while still being able to reflect the relationship between hidden variables of different modalities.
Additionally, using the Bernoulli variables as the inputs is also helpful in gradient descent optimization process. As the gradients from the function $T$ do not flow through the STE, which is a biased gradient estimator \cite{STE}.

\smallskip
Noticeably, a more direct way to enforce the inter-modal similarity is to maximize $I(\bxi;\bht)+I(\bxt;\bhi)$.
However, we find out that maximizing $I(\bxi;\bht)+I(\bxt;\bhi)$ results in similar performances compared with maximizing $I(\bhi;\bht)$, while requiring a higher computational cost.
% Furthermore, noticeably, 
% We leave this approach for a future study.

% \begin{equation}    
% \label{eq:mi_js_mu}
% \begin{split}
% % I(\bhi;\bht)&\propto \\
% \IJS(\bsmui;\bsmut) &= \DJS\left[p(\bsmui,\bsmut)\big\|p(\bsmui)\otimes p(\bsmut)\right] 
% % \\
%     % &\ge \sup_{T\in\mathcal{F}} \bbE_{p(\bhi, \bht)}[t] +\bbE_{p(\bhi)\otimes p(\bht)}[\log(2 - \exp(t))],
% \end{split}
% \end{equation}

%------------------------------------
\subsection{Minimizing modality gap}
\label{sec:SKL}
In the cross-modal retrieval task, besides having the representations that well capture input information and having MI between representations of different modalities maximized (the first and second requirements); it is also desirable for the gap between different modalities to be minimized (the third requirement).
% have the representations of different modalities well aligned with each other. 
In other words, binary codes from different modalities of the same pair should be as similar as possible \cite{DJSRH,UDCMH,FSH}. 
To achieve this requirement, we propose to minimize the symmetrized KL divergence between the two multivariate Bernoulli distributions (i.e., $P(\bhi|\bxi)$ and $P(\bht|\bxt)$) of the same pairs as follows:
%
% \begin{equation}
% \begin{split}
%     \DSKL\left[P(\bhi|\bxi)\|P(\bht|\bxt)\right] = &\hspace{2pt}\\
%     \DKL\left[P(\bhi|\bxi)\|P(\bht|\bxt)\right] &+ \DKL\left[P(\bht|\bxt)\|P(\bhi|\bxi)\right],
% \end{split}
% \end{equation}
\begin{align}
    \DSKL\left[P(\bhi|\bxi)\|P(\bht|\bxt)\right] = & \\
    \DKL\left[P(\bhi|\bxi)\|P(\bht|\bxt)\right] &+ \DKL\left[P(\bht|\bxt)\|P(\bhi|\bxi)\right], \nonumber
\end{align}
with the KL divergence between two multivariate Bernoulli distributions as
%\vspace{-4pt}
\begin{equation}
\footnotesize
\hspace{-3pt}\DKL\hspace{-1pt}\left[P(\bhi|\bxi)\|P(\bht|\bxt)\hspace{-1pt}\right]\hspace{-2pt}=\hspace{-3pt} \sum_{l=1}^{L}\hspace{-2pt}\left( \bsmui|_l\log\hspace{-1pt}\frac{\bsmui|_l}{\bsmut|_l}\hspace{-1pt}+\hspace{-1pt}(1\hspace{-1pt}-\hspace{-1pt}\bsmui|_l)\log\hspace{-1pt}\frac{1\hspace{-1pt}-\hspace{-1pt}\bsmui|_l}{1\hspace{-1pt}-\hspace{-1pt}\bsmut|_l}\right),\hspace{-4pt}
\end{equation}
where $\bsmui|_l$ and $\bsmut|_l$ are the $l$-th element of $\bsmui$ and $\bsmut$ respectively.

However, we found that strictly enforcing this property could result in an undesirable outcome, specifically, discarding modality-private information \cite{MIB}.
% \begin{equation}
%     I(\bxi;\bhi) = I(\bxi;\bhi|\bxt) +  I(\bxi;\bxt) -  I(\bxi;\bxt|\bhi)
% \end{equation}
In particular, considering the amount of information $\bhi$ contains which is unique to $\bxi$ and not shared by $\bxt$ (i.e., $I(\bxi; \bhi|\bxt)$ \footnote{The %expected value of the 
mutual information of $\bxi$ and $\bhi$ given $\bxt$.}), 
% Considering $\bhi$ and $\bht$ on the same domain $\mathbb{Z}$,
$I(\bxi; \bhi|\bxt)$ can be expressed as\footnote{A more detail derivation is provided in Appendix \ref{appendix:detail_upper_bound}.}:
%
% \begin{equation}
% \label{eq:private_info_upper_bound}
% \begin{split}
% I(&\bxi; \bhi|\bxt)= \bbE_{\bsxi;\bsxt\sim p(\bxi,\bxt)}\bbE_{\bsh\sim P_{\theta_i}\hspace{-1pt}(\bhi|\bxi)}\hspace{-2pt}\left[\log \frac{P_{\theta_i}\hspace{-1pt}(\bhi = \bsh|\bxi = \bsxi)}{P_{\theta_i}\hspace{-1pt}(\bhi = \bsh|\bxt = \bsxt)}\right]\\
% &= \bbE_{\bsxi;\bsxt\sim p(\bxi,\bxt)}\bbE_{\bsh\sim P_{\theta_i}\hspace{-1pt}(\bhi|\bxi)}\hspace{-2pt}\left[\log \frac{P_{\theta_i}\hspace{-1pt}(\bhi = \bsh|\bxi = \bsxi)}{P_{\theta_t}\hspace{-1pt}(\bht = \bsh|\bxt = \bsxt)}\right]\\
% %
% &\qquad+ \bbE_{\bsxi;\bsxt\sim p(\bxi,\bxt)}\bbE_{\bsh\sim P_{\theta_i}\hspace{-1pt}(\bhi|\bxi)}\hspace{-2pt}\left[\log \frac{P_{\theta_t}\hspace{-1pt}(\bht = \bsh|\bxt = \bsxt)}{P_{\theta_i}\hspace{-1pt}(\bhi = \bsh|\bxt = \bsxt)}\right]\\
% %
% &= \DKL\left[P_{\theta_i}\hspace{-1pt}(\bhi|\bxi)||P_{\theta_t}\hspace{-1pt}(\bht|\bxt)\right] - \DKL\left[P_{\theta_i}\hspace{-1pt}(\bht|\bxi)||P_{\theta_t}\hspace{-1pt}(\bht|\bxt)\right]\\
% &\le \DKL\left[P_{\theta_i}\hspace{-1pt}(\bhi|\bxi)||P_{\theta_t}\hspace{-1pt}(\bht|\bxt)\right].
% \end{split}
% % \vspace{-10pt}
% \end{equation}
{\small
\begin{align}
\label{eq:private_info_upper_bound}
I(&\bxi; \bhi|\bxt)\\
&= \bbE_{\bsxi;\bsxt\sim p(\bxi,\bxt)}\bbE_{\bsh\sim P_{\theta_i}\hspace{-1pt}(\bhi|\bxi)}\hspace{-2pt}\left[\log \frac{P_{\theta_i}\hspace{-1pt}(\bhi = \bsh|\bxi = \bsxi)}{P_{\theta_i}\hspace{-1pt}(\bhi = \bsh|\bxt = \bsxt)}\right] \nonumber \\
&= \bbE_{\bsxi;\bsxt\sim p(\bxi,\bxt)}\bbE_{\bsh\sim P_{\theta_i}\hspace{-1pt}(\bhi|\bxi)}\hspace{-2pt}\left[\log \frac{P_{\theta_i}\hspace{-1pt}(\bhi = \bsh|\bxi = \bsxi)}{P_{\theta_t}\hspace{-1pt}(\bht = \bsh|\bxt = \bsxt)}\right] \nonumber \\
&\qquad+ \bbE_{\bsxi;\bsxt\sim p(\bxi,\bxt)}\bbE_{\bsh\sim P_{\theta_i}\hspace{-1pt}(\bhi|\bxi)}\hspace{-2pt}\left[\log \frac{P_{\theta_t}\hspace{-1pt}(\bht = \bsh|\bxt = \bsxt)}{P_{\theta_i}\hspace{-1pt}(\bhi = \bsh|\bxt = \bsxt)}\right] \nonumber \\
&= \DKL\left[P_{\theta_i}\hspace{-1pt}(\bhi|\bxi)||P_{\theta_t}\hspace{-1pt}(\bht|\bxt)\right] - \DKL\left[P_{\theta_i}\hspace{-1pt}(\bht|\bxi)||P_{\theta_t}\hspace{-1pt}(\bht|\bxt)\right] \nonumber \\
&\le \DKL\left[P_{\theta_i}\hspace{-1pt}(\bhi|\bxi)||P_{\theta_t}\hspace{-1pt}(\bht|\bxt)\right]. \nonumber
\end{align}
}
Analogously, 
%\vspace{-3pt}
\begin{equation}
\label{eq:private_info_upper_bound1}
    I(\bxt; \bht|\bxi) \le \DKL\left[P _{\theta_t}\hspace{-1pt}(\bht|\bxt)||P_{\theta_i}\hspace{-1pt}(\bhi|\bxi)\right].
%\vspace{-3pt}
\end{equation}
Note that the upper bounds in Eq. \eqref{eq:private_info_upper_bound}, \eqref{eq:private_info_upper_bound1} are tight as the distributions of two representations coincide. Consequently, we can obtain
\begin{equation}
% \begin{split}
% I_\theta(\bxi; \bhi|\bxt) + I(\bxt; \bht|\bxi) \le & \DKL\left[P(\bhi|\bxi)||P(\bht|\bxt)\right] \\ 
% & +  \DKL\left[P (\bht|\bxt)||P(\bhi|\bxi)\right]    
% \end{split}
\label{eq:SKL_bound}
I(\bxi; \bhi|\bxt) + I(\bxt; \bht|\bxi) \le \DSKL\left[P (\bht|\bxt)||P(\bhi|\bxi)\right].
\end{equation}

From equation \eqref{eq:SKL_bound}, we can see that minimizing the difference between the binary representations of different modalities
% (to achieve consistent binary representations for pairs of samples)
(to minimize the modality gap in binary representation spaces)
would also result in discarding the modality-private information (i.e., minimizing $I(\bxi; \bhi|\bxt)$ and $I(\bxt; \bht|\bxi)$). Equivalently, the binary representations become less representative (less informative) for the inputs. 
However, having inconsistent representations for pairs of image and text samples, on the other hand, inevitably deteriorates the cross-modal retrieval performance. Therefore, it is important to appropriately weight the symmetrized KL divergence to balance the modality gap and modality-private information loss.
We will further empirically analyze this problem in the experiment section (Section \ref{sec:exp_SKL_effect}). 
%It is als worth noting that in \cite{MIB}, the authors aimed to minimize $\DSKL\left[P (\bht|\bxt)||P(\bhi|\bxi)\right]$ as much as possible to have robust representations.

\subsection{Additional properties of good hash code: Independence and Balance}
The encoded binaries in hashing algorithms are in general short in length.
To maximize hash code representative capability, 
% it is necessary to de-correlate each bit and balance the quantity of 1 and −1 in a code vector. 
we  additionally  include  the  \textit{independent}  and \textit{balancing}  regularizers  on  the  binary  codes,  i.e.,  different  bits  in the binary codes are independent to each other and each bit has 50\% chance of being $0$ or $1$, respectively \cite{spectral_hashing,SSH}. The independence property is to minimize redundant information captured in hash codes,
% ensure hash codes do not capture redundant information, 
and the balance property is to ensure hash codes contain a maximum amount of information \cite{SSH}.

\textbf{Independence:} To enhance the independence between hash bits, we aim to minimize the Total Correlation (TC) \cite{TC}, which is a popular measure of dependence for multiple random variables (i.e., multiple Bernoulli variables of multiple hash bits in our case) $TC(\bz)\coloneqq\DKL\left[q(\bz)\|q(\tilde{\bz})\right],$
% \begin{equation}
%     TC(z)\coloneqq\DKL\left[q(z)\|q(\tilde{z})\right],
% \end{equation}
where $q(\tilde{\bz}) \coloneqq \prod_{j=1}^L q(z_j)$.
However, the TC is intractable since both $q(\bz)$ and $q(\tilde{\bz})$ involve mixtures with an exponential number of components. Fortunately, being able to access to samples from both $q(\bz)$ and $q(\tilde{\bz})$ distributions\footnote{The sampling for the distribution $q(\tilde{\bz})$ can be obtained by randomly permuting across a mini-batch of samples from $q(\bz)$ for each dimension.} allows us to minimise their KL divergence using the density-ratio trick \cite{Nguyen_2010} as illustrated in \cite{factorVAE} as follows:
\begin{equation}
%\vspace{-5pt}
% \hspace{-7pt}\mathcal{L}_{\text{ind}}\hspace{-1pt}=\hspace{-1pt}
TC(\bz)=\bbE_{q(\bz)}\hspace{-3pt}\left[\log\frac{q(\bz)}{q(\tilde{\bz})}\right]
    \approx  \bbE_{q(\bz)}\hspace{-3pt}\left[\log\frac{D(\bz)}{1-D(\bz)}\right],
\end{equation}
in which the classifier $D\in[0,1]$ is jointly trained to classify between samples from $q(\bz)$ and $q(\tilde{\bz})$; and the classifier outputs the probability $D(\bz)$ that the input is a sample from $q(\bz)$ rather than from $q(\tilde{\bz})$.
% \begin{equation}
%     \mathcal{L}_{\text{ind}} = \bbE_{P(\bhi|\bxi)}\hspace{-3pt}\left[\log\frac{D^i(\bhi)}{1-D^i(\bhi)}\right] + \bbE_{P(\bht|\bxt)}\hspace{-3pt}\left[\log\frac{D^t(\bht)}{1-D^t(\bht)}\right]
% \end{equation}
Similar to the function $T$ for estimating MI lower-bound discussed in Section \ref{sec:IJS}, we also use the Bernoulli variables as the input for the classifier $D(\cdot)$.
\begin{equation}
    \mathcal{L}_{\text{ind}} = \bbE_{q(\bsmui)}\hspace{-3pt}\left[\log\frac{D^i(\bsmui)}{1-D^i(\bsmui)}\right] + \bbE_{q(\bsmut)}\hspace{-3pt}\left[\log\frac{D^t(\bsmut)}{1-D^t(\bsmut)}\right]
\end{equation}

\textbf{Balance:} To obtain balanced hash codes, we regularize the encoders such that the averaged probabilities (over the training set) for a bit to be $0$ or $1$ are equal and equal to $50\%$. Equivalently, we have
%\vspace{-5pt}
\begin{equation}
    \mathcal{L}_{\text{bal}}=\sum_{l=1}^L\bigg|\frac{1}{N}\sum_{j=1}^N\left(\bsmui_j|_l-0.5\right)\bigg|+\bigg|\frac{1}{N}\sum_{j=1}^N\left(\bsmut_j|_l-0.5\right)\bigg|,
%\vspace{-3pt}
\end{equation}
where $|\cdot|$ is the absolute function. Note that $\mathcal{L}_{\text{bal}}$ can be minimized in mini-batch manner.

\subsection{Final objective function and reference stage}
In summary, the final objective function of our proposed method is defined as follows:
\begin{equation}
\begin{split}
    \max~~ I(\bxi;\bhi) &+ I(\bxt;\bht) + \lambda_1 I(\bhi;\bht) \\
    &- \lambda_2 \DSKL(\bhi;\bht) - \lambda_3 \mathcal{L}_{\text{ind}} - \lambda_4 \mathcal{L}_{\text{bal}},
\end{split}
%\vspace{-3pt}
\end{equation}
in which $\lambda_1, \lambda_2,\lambda_3,$ and $\lambda_4$ are hyper-parameters.

\smallskip

For reference, it is undesirable to have different binary code for a query sample under different retrieval runs; hence, we obtain the deterministic binary codes by simply applying a threshold function on the Bernoulli variables, i.e., $\bh = \sign(\bsmu-0.5)$.

%=====================================================================

\section{Experiment}
In this section, we conduct a wide range of experiments to validate our proposed method on three standard benchmark datasets for the cross-model retrieval task, i.e., {MIR-Flickr25k}
% \footnote{https://press.liacs.nl/mirflickr/} 
\cite{MIRFlickr25k},
% {IAPR-TC12}\footnote{https://www.imageclef.org/photodata}, and
{NUS-WIDE} \cite{nuswide},
% \footnote{https://lms.comp.nus.edu.sg/research/NUS-WIDE.htm} 
and MS-COCO \cite{mscoco}.

\subsection{Experiment setting}
\label{sec:exp_settings}
\textbf{Datasets:} 
% The \textbf{Wiki} dataset consists of 2,866 multimedia documents in 10
% categories from Wikipedia. Each document serves as an instance containing one image and text with at least 70 words.
% A hand-crafted 128-dimensional SIFT feature vector is also provided for each image, while each text is accompanied
% with a 10-dimensional topic vector generated by the Latent
% Dirichlet Allocation (LDA) model.
%
The \textbf{MIR-Flickr25K} dataset \cite{MIRFlickr25k} is collected from Flickr website, which contains 25,000 image-text pairs together with 24 provided labels. The texts are represented as 1386-dimensional tagging vectors. Additionally, we remove the pairs whose texts do not contain any tag in the 1,386 common tags results. As a result, 20,015 pairs are preserved. Following \cite{UDCMH,DJSRH}, we randomly sample 2,000 instances for the query set while the remaining instances are used as the database. Additionally, 5,000 instances are randomly sampled from the database to form the training set.

The \textbf{NUS-WIDE} dataset \cite{nuswide} is a multi-label image dataset crawled from Flickr, which contains 296,648 images with associated tags. Each image-tag pair is annotated with one or more labels from 81 concepts. 
In this dataset, each text is represented by a 1,000-dimension preprocessed BOW feature. 
Following the common practice \cite{FSH,DDCMH,DJSRH}, we select image-tag pairs which have at least one label belonging to the top 10 most frequent concepts and the corresponding 186,577 annotated instances are preserved.
We randomly sample 2,000 instances as queries. The remaining instances are used as the database, and 5,000 instances are randomly sampled from the database to form the training set.
% We randomly choose 200 image-tag pairs per semantic label as the query set of total 2,000 pairs, and the remaining as the database. 
% We then randomly sample 20,000 pairs from the database to form the training set.

The \textbf{MS-COCO}-2017 consists of 118,287 training images and 5,000 validation images. Each image includes at least five sentences annotations (captions). We randomly select one sentence and use the pretrained BERT model \cite{BERT} to extract the sentence embedding as the text representations.
Following \cite{CYC-DGH}, we use the provided 80 image segmentation categories as ground truth labels for the image-sentence pairs. We use the validation set as the query set. By removing image-sentence pairs that have no category information, we obtain 117,266 database samples and 4,952 query samples. Similar to the MIR-Flickr25K and NUS-WIDE datasets, we randomly sample 5,000 instances from the database for training.

% \smallskip
For images of all datasets, we extract FC7 features from the PyTorch pretrained AlexNet network \cite{alexnet}, and then apply PCA to compress to 1024-dimension.
% Wiki is officially split into the database and the query set with 2,173 and 693 instances respectively.

% \smallskip
% \subsubsection{Evaluation Metrics:}  
\textbf{Evaluation Metrics:}
The evaluations are presented in both cross-modal retrieval tasks (i.e., Img $\to$ Txt, Txt $\to$ Img) and single-modal retrieval tasks (i.e., Img $\to$ Img, Txt $\to$ Txt); in which images (Img)/texts (Txt) are used as queries to retrieve image/text database samples accordingly.
The quantitative performance is evaluated by the standard evaluation metrics: \textit{(i)} mean Average Precision of top 1000 returned samples (\textbf{\textit{mAP@1k}}) and \textit{(ii)} precision curve at top-$K$ retrieved images (\textbf{\textit{Prec@K}}).
% \mAP and \textbf{\textit{Pre@r2}} are widely used metrics for evaluating the accuracy of hashing. 
The image-text pairs are considered to be similar if they share at least one common label. Otherwise, they are considered to be dissimilar. 
% To minimize the bias due to the random query and training samples, we conduct all the experiments 5 times with 5 different random query and training sets and report the average values.

% \subsubsection{Implementation Details:}
\textbf{Implementation Details:}
% For fair comparison, we use the extracted features, which are used as the input for compared methods, to construct anchor graphs. 
% For anchor graph, we set $P\eq 500, k\eq 3$ and $k_a\eq 2$. We learn $\sigma$ by the mean squared distances to $k$-nearest neighborhoods. In addition, we empirically set $\lambda_1\eq\lambda_1\eq 1, \gamma_1\eq\gamma_2\eq 100$, and $\alpha\eq 1$ by cross validation. More details about CNN models are presented in Supplementary. 
% % The above setting is used in our experiments unless stated otherwise.
%
% We clamp the output of encoder before the sigmoid activation to be within $[-5, 5$
%
%
Both encoder and decoder consist of multi-layer perceptrons (MLP) of two hidden ReLU units of size 1,024. 
The critic $T$ for $\IJS$ estimator \eqref{eq:mi_js} is a separable function $f(x; y) = \phi^i(x)^\top\phi^t(y)$, where $\phi^i(\cdot)$ and $\phi^t(\cdot)$ are MLPs with two hidden layers of size 512 and Leaky-ReLU activations. 
The classifier $D$ to estimate $TC$ also  consists of a MLP of two hidden Leaky-ReLU units of size 512.

Additionally, we employ the SGD optimizer with mini-batch size of $128$, momentum of $0.9$ and weight decay of $0.0001$. The learning rate is set as $0.01$ for the encoders, the critic $T$ and the classifier $D$ , and set as $0.001$ for the decoders. The hyper-parameters $\lambda_1, \lambda_2, \lambda_3$ and $\lambda_4$ are empirically set by cross validation as $1.5$, $1$, $0.25$, and $0.01$ respectively for MIR-Flickr25k and NUS-WIDE datasets and set as $4$, $1.5$, $0.25$, and $0.01$ respectively for MS-COCO dataset.
% For fair comparison, the same input features as described in the dataset section are used for all compared methods and our method.

%\vspace{-3pt}
\subsection{Ablation Study and Parameter Analysis}
\subsubsection{The necessity of explicitly maximizing the mutual information between hash codes of different modalities.}

\begin{figure}[t]
\centering
\caption{The {\textit{mAP@1K}} curves as the weight of $I(\bh^i,\bh^t)$ varies on MIR-Flickr25k and NUS-WIDE datasets using 32-bit hash codes.}
\label{fig:map_by_mi}
\begin{subfigure}[b]{0.22\textwidth}
\centering
\caption{MIR-Flickr25k}
\includegraphics[width=\textwidth]{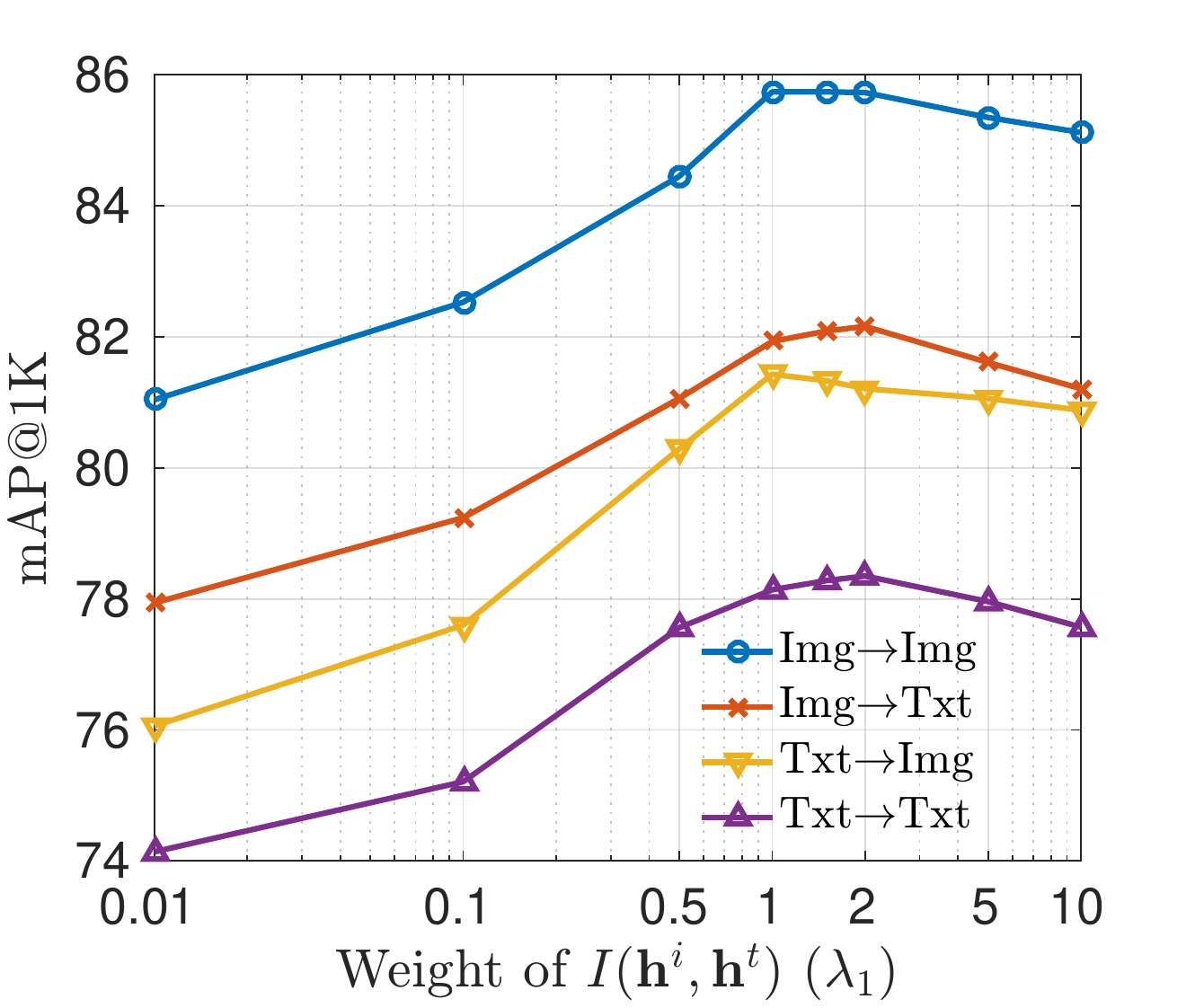}
\end{subfigure}
\begin{subfigure}[b]{0.22\textwidth}
\centering
\caption{NUS-WIDE}
\includegraphics[width=\textwidth]{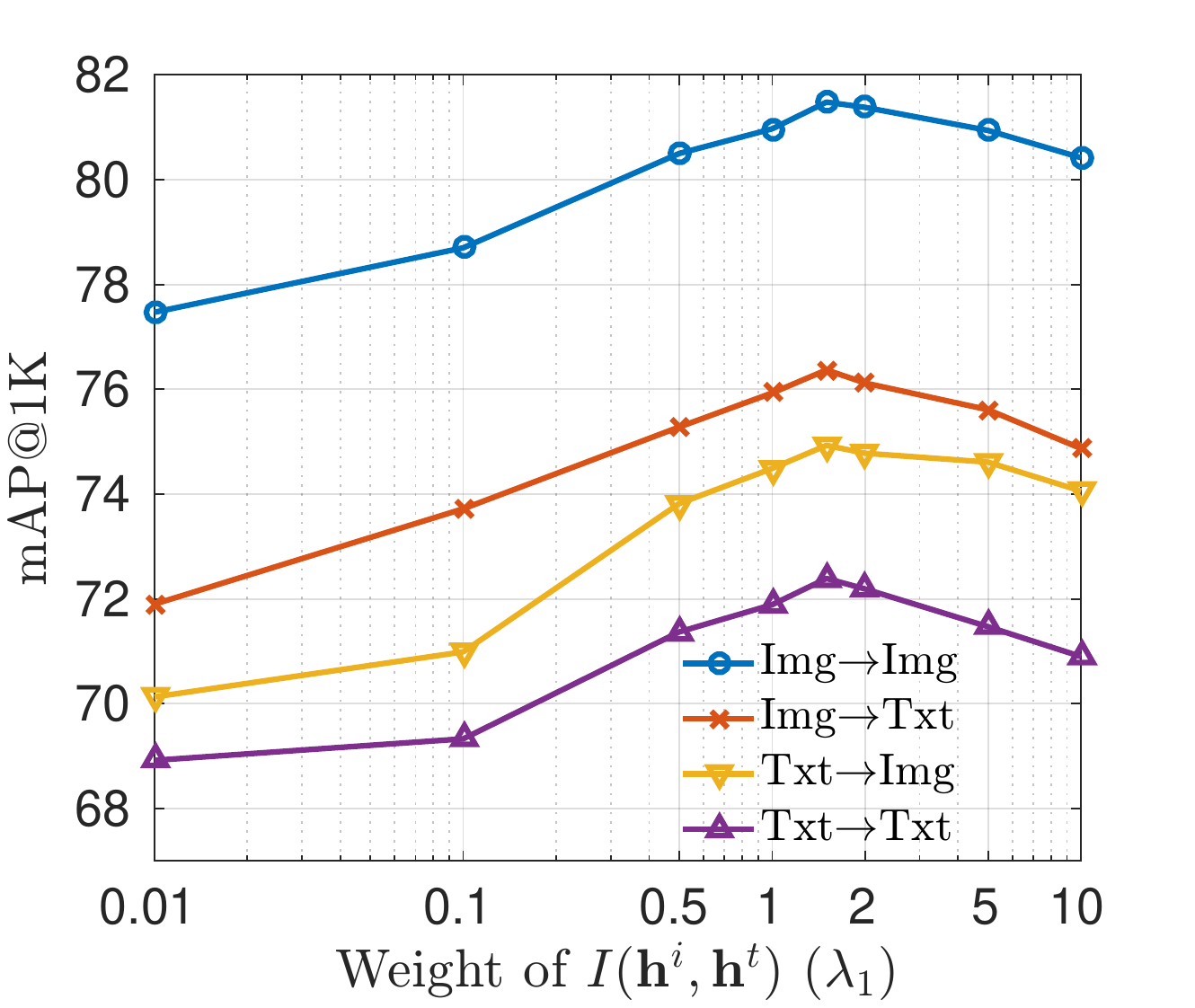}
\end{subfigure}
%\vspace{-10pt}
\end{figure}

In this section, we conduct experiments on MIR-Flickr25k and NUS-WIDE datasets using 32-bit hash codes with various values of $I(\bh^i,\bh^t)$ weight (i.e., $\lambda_1$). The experimental results in term of \textit{mAP@1K} are presented in Figure \ref{fig:map_by_mi}. 
As can be seen, when using a very small weight for $I(\bh^i,\bh^t)$ (i.e., $\lambda_1 \le 0.1$), the retrieval performance is significantly lower for all four retrieval tasks in compared with larger $I(\bh^i,\bh^t)$ weights (i.e., $\lambda_1 \ge 1$).
With a reasonable large weight for $I(\bh^i,\bh^t)$, the model is enforced to retain the information that is shared across modalities. The information that is shared among different modalities information is generally more useful for both the cross-modal and single-modal retrieval tasks; as, intuitively, this type of information is more likely to contain the ground-truth information. The experimental results confirm the importance and necessity of explicitly maximizing the mutual information between hash codes of different modalities.
Besides, at a too large $I(\bhi,\bht)$ weight (i.e., $\lambda_1 \ge 5$), we also observe small decreases in retrieval performance. This fact is also understandable as the model pays less attention on maximizing $I(\bx^i,\bhi)$ and $I(\bx^t,\bht)$, which results in less informative hash codes. 
\begin{figure}[t]
\centering
\caption{Histogram of $\bsmui$ when using $T(\bsmui,\bsmut)$ and $T(\bshi,\bsht)$ (single pair of binary samples) for the MI lower bound estimator. The experiment is conducted on MIR-Flicrk25k with 32-bit hash codes.  We observe similar histograms for $\bsmut$.}
\label{fig:hist}
\begin{subfigure}[b]{0.21\textwidth}
\centering
\caption{$T(\bsmui,\bsmut)$}
\includegraphics[width=\textwidth]{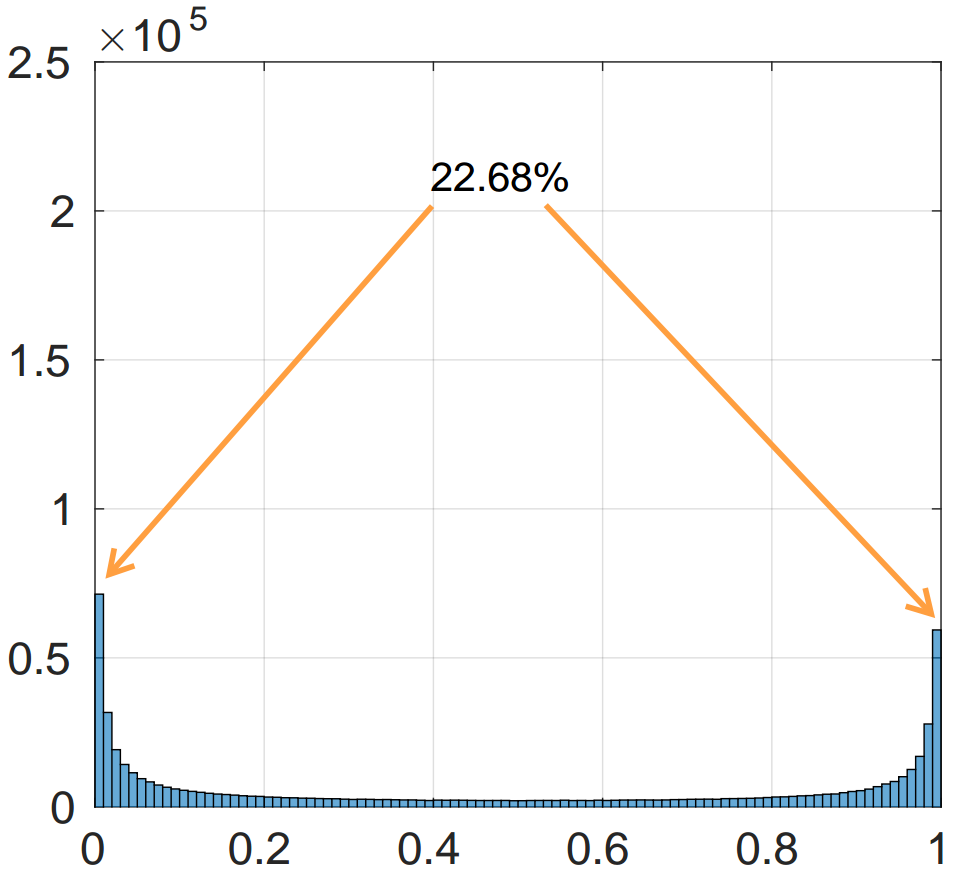}
\end{subfigure}
$\quad$
\begin{subfigure}[b]{0.21\textwidth}
\centering
\caption{$T(\bshi,\bsht)$}
\includegraphics[width=\textwidth]{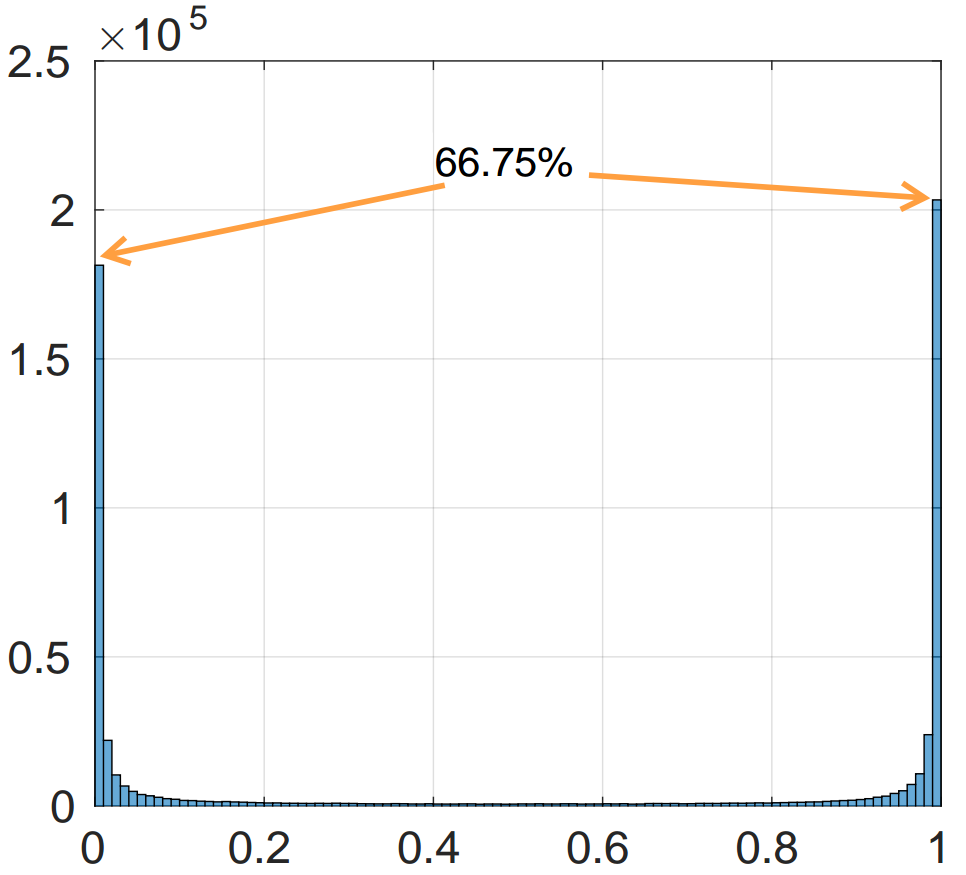}
\end{subfigure}
%\vspace{-10pt}
\end{figure}

\begin{table}[t]
\small
\def\arraystretch{1.1}
\setlength{\tabcolsep}{5pt}
\centering
\caption{Cross-modal retrieval performance (\textit{mAP@1K} (\%)) with different input settings for function $T(\cdot)$ in estimating $I(\bhi,\bht)$ lower bound. The experiment is conducted on MIR-Flicrk25k with 32 bit hash codes.}
\label{tb:multi_sample}
\resizebox{1.0\columnwidth}{!}
{
\begin{tabular}{|c|c|c|c|c|c|c|c|}
\hline
\multirow{2}{*}{Task} & \multirow{2}{*}{$T(\bsmui,\bsmut)$} & \multicolumn{4}{c|}{\# of pairs of bin. samples for $T(\bshi,\bsht)$}  \\ \cline{3-6} 
            & & 1 & 5 & 10 & 20  \\ \hline
Img$\to$Txt & \textbf{81.93}  & \hspace{4pt}80.24\hspace{4pt} & \hspace{4pt}81.27\hspace{4pt} & \hspace{4pt}81.75\hspace{4pt} & 81.74 \\ \hline
Txt$\to$Img & \textbf{81.43}  & \hspace{4pt}79.41\hspace{4pt} & \hspace{4pt}80.86\hspace{4pt} & \hspace{4pt}81.22\hspace{4pt} & 81.29 \\ \hline
\end{tabular}
}
%\vspace{-10pt}
\end{table}

%\vspace{-5pt}
\subsubsection{The benefit of using Bernoulli variables as the input of function $T$ in estimating $I(\bhi,\bht)$ lower bound} 
In this section, we conduct experiments on MIR-Flickr25 dataset with $L=32$ bits to validate the benefit of using Bernoulli variables as the input of function $T$ in estimating $I(\bhi,\bht)$ lower bound in comparison with using multiple pairs of binary samples as the input. The retrieval performance is presented in Table \ref{tb:multi_sample}. We also present the histograms of $\bsmui$ when using $T(\bsmui,\bsmut)$ and $T(\bshi,\bsht)$ (with single sample) for the MI lower bound estimator in Figure \ref{fig:hist}. When using single pairs of binary samples as the input for the function $T(\bshi,\bsht)$, we can observe that the majority values of $\bsmui$ become very small or very large. Approximate $67\%$ of $\bsmui$ are in $[0, 0.01]$ or $[0.99, 1]$ (i.e., to be $0$ or $1$ respectively with $99\%$ confident), in comparison with about $22\%$ when using $T(\bsmui,\bsmut)$. This effect significantly affects the retrieval performance. When using multiple binary samples, the performance improves. However, the best performance is achieved when using Bernoulli variables. As this not only helps to eliminate input noise; but it also helps the gradients to not propagate through the biased gradient estimator STE, which introduces noise in gradients.
Furthermore, we note that using $n$ binary samples would require approximately $n$-times computational cost.

%\vspace{-3pt}
\subsubsection{Effect the symmetrized KL divergence}
\label{sec:exp_SKL_effect}

\begin{figure}[t]
\centering
\caption{The {\textit{mAP@1K}} (\%) curves as the weight of $\DSKL$ varies on MIR-Flickr25k and NUS-WIDE using 32-bit hash codes.}
\label{fig:map_by_skl}
\begin{subfigure}[b]{0.22\textwidth}
\centering
\caption{MIR-Flickr25k}
\includegraphics[width=\textwidth]{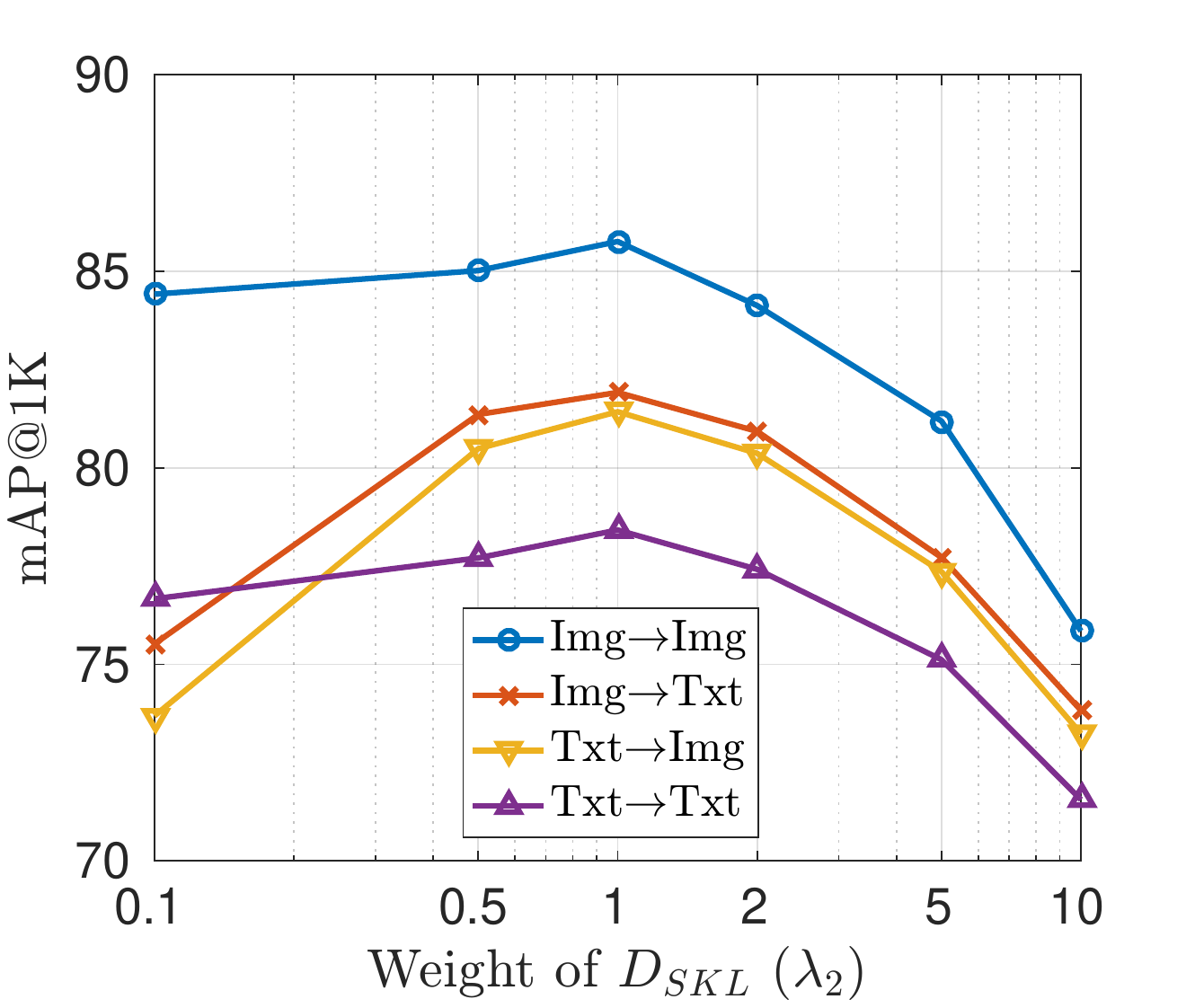}
\end{subfigure}
\begin{subfigure}[b]{0.22\textwidth}
\centering
\caption{NUS-WIDE}
\includegraphics[width=\textwidth]{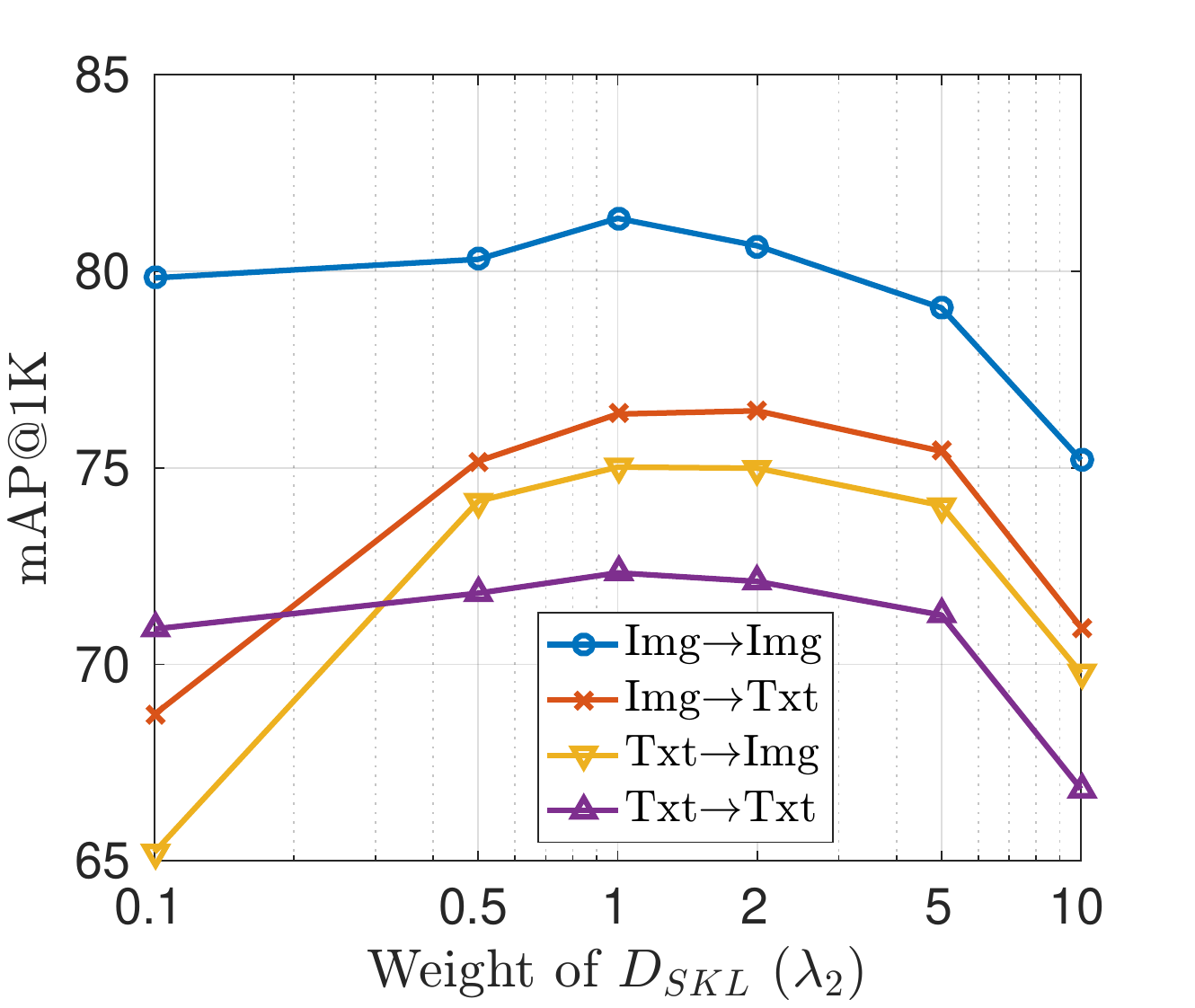}
\end{subfigure}
%\vspace{-10pt}
\end{figure}

\begin{table}[t]
\def\arraystretch{1.1}
\setlength{\tabcolsep}{3.5pt}
\centering
\caption{The TC, Corr MSE, and cross-modal retrieval performance (\textit{mAP@1K} (\%)) for MIR-Flickr25k dataset at different code lengths. \cmark and \xmark ~indicate the hash codes are learned \textit{with} and \textit{without} $\mathcal{L}_{\text{ind}}$, respectively.}
\label{tb:TC}
% \vspace{5pt}
\resizebox{1.0\columnwidth}{!}
{
\begin{tabular}{|c|c|c|c|c|c|c|c|}
\hline
 & & \multicolumn{2}{c|}{16} & \multicolumn{2}{c|}{32} & \multicolumn{2}{c|}{48} \\ \cline{3-8}
 & & \cmark & \xmark & \cmark & \xmark & \cmark & \xmark   \\ \hline
\multirow{2}{*}{TC} & Img & 2.327 & 4.491 & 6.142 & 6.418 & 6.514 & 6.718  \\
& Txt & 2.174 & 4.288 & 6.015 & 6.424 & 6.425 & 6.645  \\ \hline
\multirow{2}{*}{Corr MSE} & Img & 0.037 & 0.068 & 0.040 & 0.092 & 0.051 & 0.078  \\
& Txt & 0.045 & 0.079 & 0.047 & 0.110 & 0.064 & 0.095  \\ \hline
\multirow{2}{*}{\textit{mAP@1K}} & Img$\to$Txt & 80.68 & 80.39 & 81.93 & 81.38 & 82.92 & 82.16  \\
& Txt$\to$Img & 79.77 & 79.76 & 81.43 & 81.14 & 82.18 & 81.28  \\
\hline
\end{tabular}
}
%\vspace{-7pt}
\end{table}

%==========================================================================

\begin{table*}[t]
\small
%\large
\centering
\def\arraystretch{1.1}
\setlength{\tabcolsep}{8pt}
\caption{Comparison with state-of-the-art methods using \textit{mAP@1k} (\%) on three benchmark datasets. 
% The performances of SOTA-FSH are cited from \cite{FSH}.
}
\label{tb:img_txt}
% \vspace{5pt}
\begin{tabular}{|c|c|c|c|c!{\vrule width 1pt}c|c|c!{\vrule width 1pt}c|c|c|}
\hline
\multirow{2}{*}{Task} & \multirow{2}{*}{Method} & \multicolumn{3}{c!{\vrule width 1pt}}{MIR-Flickr25k} & \multicolumn{3}{c!{\vrule width 1pt}}{NUS-WIDE} & \multicolumn{3}{c|}{MS-COCO} \\ \cline{3-11}
% & & \multicolumn{3}{c|}{\textit{mAP}} & \multicolumn{3}{c|}{\textit{mAP}} & \multicolumn{3}{c|}{\textit{mAP}} \\ \cline{3-11}
&  & 16 & 32 & 48 & 16 & 32 & 48 & 16 & 32 & 48  \\
\hline 

% \hline\hline
\multirow{6}{*}{\rotatebox{90}{Img$\to$Txt}} 
& CVH \cite{CVH}  & 68.18 & 66.95 & 66.32 & 56.43 & 57.16 & 57.47 & 61.49 & 62.15 & 60.06 \\ \cline{2-11}
& PDH \cite{PDH} & 78.16 & 79.62 & 81.10 & 70.98 & 74.21 & 75.13 & 61.66 & 65.60 & 67.27 \\ \cline{2-11}
& CMFH \cite{CMFH} & 77.97 & 78.69 & 78.50 & 69.13 & 70.96 & 71.18 & 59.21 & 64.62 & 66.55 \\ \cline{2-11}
& ACQ \cite{ACQ} & 76.16 & 76.50 & 76.93 & 67.35 & 70.05 & 70.88 & 60.66 & 63.24 & 65.66 \\ \cline{2-11}
& FSH \cite{FSH} & 77.55 & 79.36 & 80.52 & 69.45 & 70.48 & 72.72 & 62.75 & 66.24 & 69.04 \\ \cline{2-11}
& {DJSRH} \cite{DJSRH} & 79.05 & 79.53 & 81.66 & 71.23 & 74.86 & 76.52 & 59.39 & 67.32 & 68.50 \\ \cline{2-11}
& \textbf{CMIMH} & \textbf{80.68} & \textbf{81.93} & \textbf{82.92} & \textbf{73.92} & \textbf{76.37} & \textbf{77.21} & \textbf{65.32} & \textbf{69.21} & \textbf{70.20} \\ \hline

% \hline\hline
\multirow{6}{*}{\rotatebox{90}{Txt$\to$Img}} 
& CVH & 68.08 & 66.89 & 66.40 & 57.40 & 58.30 & 58.51 & 62.45 & 63.46 & 61.22 \\ \cline{2-11}
& PDH & 76.79 & 78.64 & 79.22 & 69.61 & 72.24 & 73.88 & 63.24 & 67.69 & 69.66 \\ \cline{2-11}
& CMFH & 76.81 & 76.83 & 77.36 & 66.98 & 69.14 & 70.32 & 60.20 & 66.06 & 68.39 \\ \cline{2-11}
& ACQ & 74.46 & 75.22 & 75.39 & 65.53 & 68.22 & 69.54 & 61.83 & 64.44 & 66.96 \\ \cline{2-11}
& FSH & 75.10 & 77.10 & 78.47 & 67.59 & 69.03 & 70.25 & 65.05 & 69.07 & 71.48 \\ \cline{2-11}
& {DJSRH}  & 77.44 & 78.65 & 80.10 & 68.18 & 73.29 & 74.72 & 56.19 & 67.95 & 71.11 \\ \cline{2-11}
& \textbf{CMIMH} & \textbf{79.77} & \textbf{81.43} &  \textbf{82.18} & \textbf{72.75} & \textbf{75.02} & \textbf{75.68} & \textbf{66.08} & \textbf{70.35} & \textbf{72.21} \\ \hline
% \end{tabular}
% \end{table*}

% % =====================================================================================
% \begin{table*}[t]
% \small
% \centering
% \setlength{\tabcolsep}{4.2pt}
% \caption{Comparison results using \textit{mAP@1k} (\%) on three benchmark datasets.}
% \label{tb:img_img_txt_txt}
% \begin{tabular}{|c|c|c|c|c|c|c|c|c|c|c|}
% \hline
% \multirow{2}{*}{Task} & \multirow{2}{*}{Method} & \multicolumn{3}{c|}{MIR-Flickr25k} & \multicolumn{3}{c|}{IAPR-TC12} & \multicolumn{3}{c|}{NUS-WIDE} \\ \cline{3-11}
% % & & \multicolumn{3}{c|}{\textit{mAP}} & \multicolumn{3}{c|}{\textit{mAP}} & \multicolumn{3}{c|}{\textit{mAP}} \\ \cline{3-11}
% &  & 16 & 32 & 48 & 16 & 32 & 48 & 16 & 32 & 48  \\
% \hline \hline
\multirow{6}{*}{\rotatebox{90}{Img$\to$Img}} 
% & ITQ \cite{ITQ} & 81.94 & 82.51 & 83.33 & 00.00 & 00.00 & 00.00 & 00.00 & 00.00 & 00.00 \\ \cline{2-11}
& CVH & 69.49 & 68.36 & 67.76 & 59.64 & 60.41 & 61.25 & 60.66 & 61.97 & 60.94 \\ \cline{2-11}
& PDH & 79.65 & 81.46 & 82.86 & 74.23 & 77.06 & 78.08 & 61.10 & 65.28 & 67.06 \\ \cline{2-11}
& CMFH & 81.45 & 82.54 & 83.13 & 75.33 & 77.72 & 78.35 & 60.31 & 65.53 & 67.78 \\ \cline{2-11}
& ACQ & 78.91 & 78.94 & 79.58 & 71.13 & 76.63 & 74.88 & 60.07 & 62.46 & 64.76 \\ \cline{2-11}
& FSH & 80.22 & 82.39 & 83.68 & 73.28 & 75.57 & 76.59 & 61.70 & 65.64 & 68.47 \\ \cline{2-11}
& {DJSRH} & 82.39 & 83.17 & 84.07 & 77.29 & 79.29 & 80.27 & 59.87 & 66.77 & 68.29 \\ \cline{2-11}
& \textbf{CMIMH} & \textbf{83.79} & \textbf{85.74} & \textbf{86.76} & \textbf{78.74} & \textbf{81.35} & \textbf{81.95} & \textbf{65.38} & \textbf{69.24} & \textbf{70.16} \\ \hline

% \hline\hline
\multirow{6}{*}{\rotatebox{90}{Txt$\to$Txt}} 
% & ITQ & 70.29 & 70.64 & 71.46 & 00.00 & 00.00 & 00.00 & 00.00 & 00.00 & 00.00 \\ \cline{2-11}
& CVH & 67.47 & 66.79 & 66.72 & 57.45 & 60.13 & 61.50 & 64.17 & 66.84 & 64.96 \\ \cline{2-11}
& PDH & 75.72 & 76.66 & 78.04 & 67.61 & 71.02 & 71.63 & 64.27 & 70.16 & 72.36 \\ \cline{2-11}
& CMFH & 74.48 & 74.92 & 75.38 & 65.92 & 68.00 & 69.64 & 62.16 & 69.09 & 71.17 \\ \cline{2-11}
& ACQ & 72.77 & 73.91 & 74.18 & 63.68 & 67.13 & 68.72 & 63.28 & 66.38 & 69.45 \\ \cline{2-11}
& FSH & 73.35 & 75.03 & 76.28 & 65.89 & 67.55 & 69.10 & 67.05 & {72.78} & \textbf{75.67} \\ \cline{2-11}
& {DJSRH} & 75.34 & 76.48 & 77.51 & 67.22 & 71.01 & 72.27 & 64.55 & 72.44 & {74.94} \\ \cline{2-11}
& \textbf{CMIMH} & \textbf{77.36} & \textbf{78.42} & \textbf{79.01} & \textbf{70.18} & \textbf{72.33} & \textbf{72.72} & \textbf{68.15} & \textbf{73.28} & {74.89} \\ \hline
\end{tabular}
%\vspace{-5pt}
\end{table*}

In Figure \ref{fig:map_by_skl}, we present the \textit{mAP@1K} curves as the weight of $\DSKL$ varies on the MIR-Flickr25k and NUS-WIDE datasets. Firstly, we can observe that too large weights for $\DSKL$ (i.e., $\lambda_2\ge 5$) have significant impacts on the retrieval performance for all four retrieval tasks. This observation is consistent with our discussion in Section \ref{sec:SKL} that too large $\DSKL$ weights will force the model to discard a large amount of modality-private information in the representations. As a result, the binary representations do not well-represent for the input data.
For too small $\DSKL$ weights (i.e., $\lambda_2<0.5$), the retrieval performance on Img $\to$ Txt and Txt $\to$ Img retrieval tasks is unsurprisingly low as the binary representations of image and text modalities are poorly aligned with each other and not suitable for the cross-modal retrieval tasks. However, different from the case of too large $\DSKL$ weights, too small $\DSKL$ weights only result in minor performance drops for the Img $\to$ Img and Txt $\to$ Txt retrieval tasks,  which means that the learned binary representations still well capture information of the input data. The small performance drops for the Img $\to$ Img and Txt $\to$ Txt retrieval tasks potentially indicate that the binary representations also capture information from the input data that does not share with the ground truth (e.g., noise). 

%\vspace{-5pt}
\subsubsection{The effectiveness of using Total Correlation (TC) as a regularizer to enhance hash bit independence}
We conduct experiments on the MIR-Flickr25k dataset with and without the independence regularizer $\mathcal{L}_{\text{ind}}$. In Table \ref{tb:TC}, we show the Mean Square Error (MSE) between the correlation matrix of the binary code of the database and the identity matrix (i.e., $\text{Corr~MSE} = \left\|\frac{1}{N}\hat{\bH}^\top \hat{\bH}-\bI_L\right\|_F^2$, where $\hat{\bH}=\{\hat{\bh}_j\}_{j=1}^N\in\{-1,1\}^{N\times L}$ is the set of $L$-bit hash codes of a dataset and $\hat{\bh}_j=2{\bh}_j-1$) together with the retrieval performance at different code lengths. As can be seen, $\mathcal{L}_{\text{ind}}$ consistently helps to reduce the Corr MSEs for both image and text modalities at various code lengths. A smaller Corr MSE indicates that the hash bits are more independence and consequently leads to higher performance.
Interestingly, we also notice that, at high code lengths, even though the classifiers can easily predict if a sample is from $q(z)$ with very high confident (i.e., high TC\footnote{The probability, that the classifier predicts a sample from $q(z)$, can be computed as $1/(1+\exp(-TC))$ (e.g., $1/(1+\exp(-6.0))\approx 99.75\%$)}), $\mathcal{L}_{\text{ind}}$ is still helpful in reducing correlation between hash bits and improving performance.

\begin{table}[t]
\def\arraystretch{1.1}
\setlength{\tabcolsep}{3.5pt}
\centering
\caption{The  cross-modal retrieval performance (\textit{mAP@1K} (\%)) for MIR-Flickr25k dataset to evaluate the effectiveness of each component in CMIMH.
\gcmark and \rxmark ~indicate the hash codes are learned \textit{with} and \textit{without} the according component, respectively. I$\to$T and T$\to$I mean Img$\to$Txt and Txt$\to$Img respectively.}
\label{tb:ablation_study}
% \vspace{5pt}
\resizebox{1.0\columnwidth}{!}
{
\begin{tabular}{|c|c|c|c|c|c|c|c|}
\hline
\multicolumn{4}{|c|}{Configuration} & \multicolumn{2}{c|}{32 bits} & \multicolumn{2}{c|}{48 bits} \\ \hline
$I(\bhi;\bht)$ & $\DSKL$ & $\mathcal{L}_{\text{ind}}$ & $\mathcal{L}_{\text{bal}}$ & I$\to$T & T$\to$I & I$\to$T & T$\to$I \\ \hline
\gcmark & \rxmark & \rxmark & \rxmark & 75.13 & 73.14 & 77.57 & 75.49 \\ \hline
\rxmark & \gcmark & \rxmark & \rxmark & 77.61 & 75.53 & 78.89 & 78.39 \\ \hline
\gcmark & \rxmark & \gcmark & \gcmark & 75.52 & 73.69 & 77.84 & 75.83 \\ \hline
\rxmark & \gcmark & \gcmark & \gcmark & 77.87 & 75.95 & 79.34 & 78.69 \\ \hline
\gcmark & \gcmark & \rxmark & \rxmark & 81.28 & 80.73 & 82.07 & 81.51 \\ \hline
\gcmark & \gcmark & \gcmark & \rxmark & 81.35 & 80.93 & 82.11 & 81.82 \\ \hline
\gcmark & \gcmark & \rxmark & \gcmark & 81.43 & 81.14 & 82.16 & 81.78 \\ \hline
\gcmark & \gcmark & \gcmark & \gcmark & \textbf{81.93} & \textbf{81.43} & \textbf{82.92} & \textbf{82.18} \\ \hline

%  & & \multicolumn{2}{c|}{16} & \multicolumn{2}{c|}{32} & \multicolumn{2}{c|}{48} \\ \cline{3-8}
%  & & \cmark & \xmark & \cmark & \xmark & \cmark & \xmark   \\ \hline
% \multirow{2}{*}{TC} & Img & 2.327 & 4.491 & 6.142 & 6.418 & 6.514 & 6.718  \\
% & Txt & 2.174 & 4.288 & 6.015 & 6.424 & 6.425 & 6.645  \\ \hline
% \multirow{2}{*}{Corr MSE} & Img & 0.037 & 0.068 & 0.040 & 0.092 & 0.051 & 0.078  \\
% & Txt & 0.045 & 0.079 & 0.047 & 0.110 & 0.064 & 0.095  \\ \hline
% \multirow{2}{*}{\textit{mAP@1K}} & Img$\to$Txt & 80.68 & 80.39 & 81.63 & 81.00 & 82.92 & 82.16  \\
% & Txt$\to$Img & 79.77 & 79.76 & 81.13 & 81.04 & 82.18 & 81.28  \\
% \hline
\end{tabular}
}
%\vspace{-7pt}
\end{table}

% \smallskip
% \smallskip
\subsubsection{A summary of effectiveness of different components}
We additionally present in Table \ref{tb:ablation_study} the  cross-modal retrieval performance (\textit{mAP@1K} (\%)) for MIR-Flickr25k dataset with 32 and 48 bit hash codes with different combinations of components in the loss function.
{
We can observe that the two terms  $I(\bhi;\bht)$ and $ \DSKL(\bhi;\bht)$ play very important roles in our proposed cross-modal hashing method. Without either $I(\bhi;\bht)$ or $ \DSKL(\bhi;\bht)$ the cross-modal retrieval performance is significantly degraded. 
The ablation study also shows that the independence and balance terms are beneficial for hashing methods. However, even without these two terms, our proposed method CMIMH still achieves very good performance.
}

%\max~~ I(\bxi;\bhi) &+ I(\bxt;\bht) + \lambda_1 I(\bhi;\bht) \\
    % &- \lambda_2 \DSKL(\bhi;\bht) - \lambda_3 \mathcal{L}_{\text{ind}} - \lambda_4 \mathcal{L}_{\text{bal}},

\begin{figure*}[t]
\centering
\caption{The {\textit{Pre@K}} (\%) curves on different datasets of 32-bit hash codes.}
\label{fig:pre@K}
\vspace{2pt}
\begin{subfigure}[b]{\textwidth}
\centering
\includegraphics[width=0.30\textwidth]{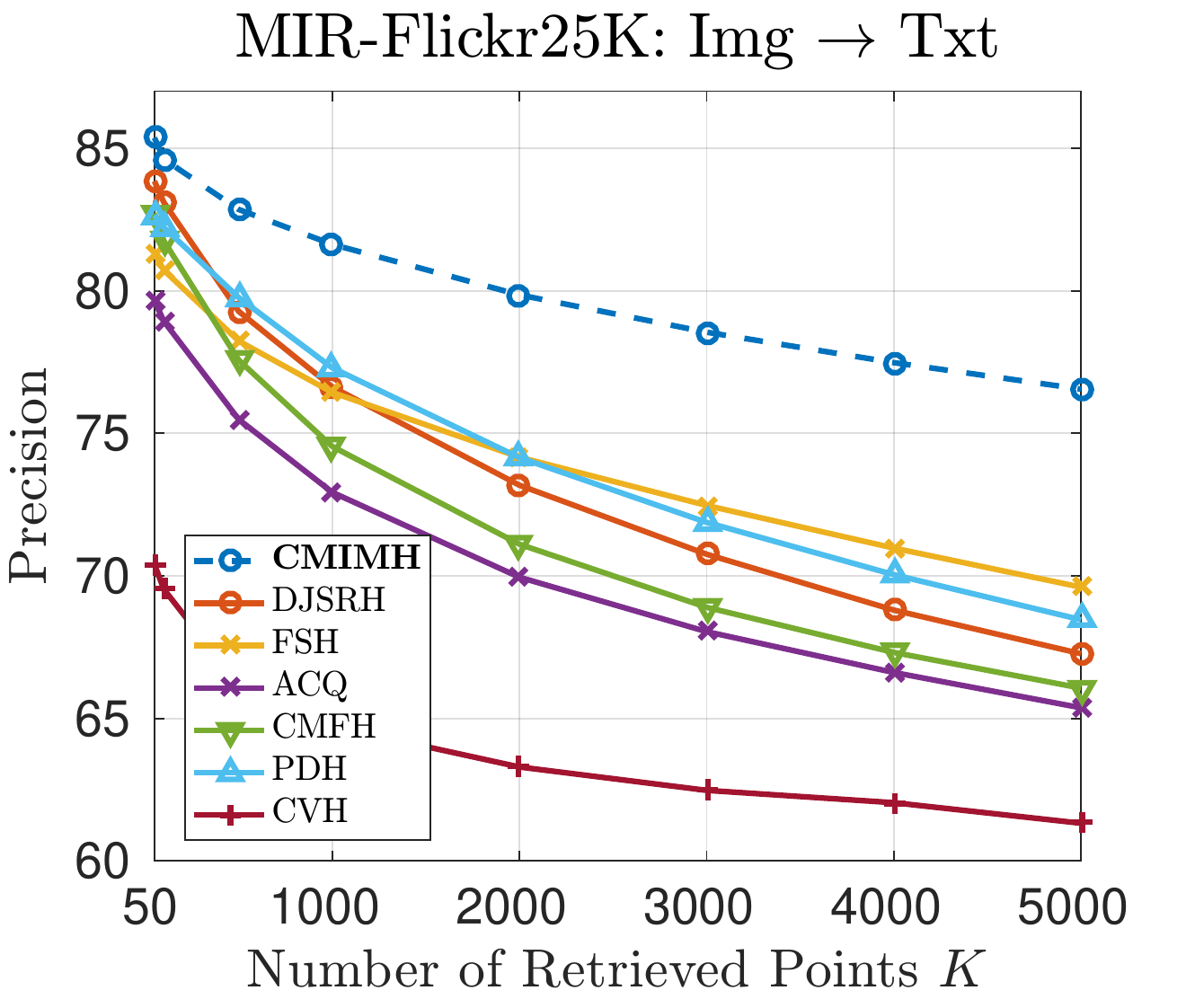}
\includegraphics[width=0.30\textwidth]{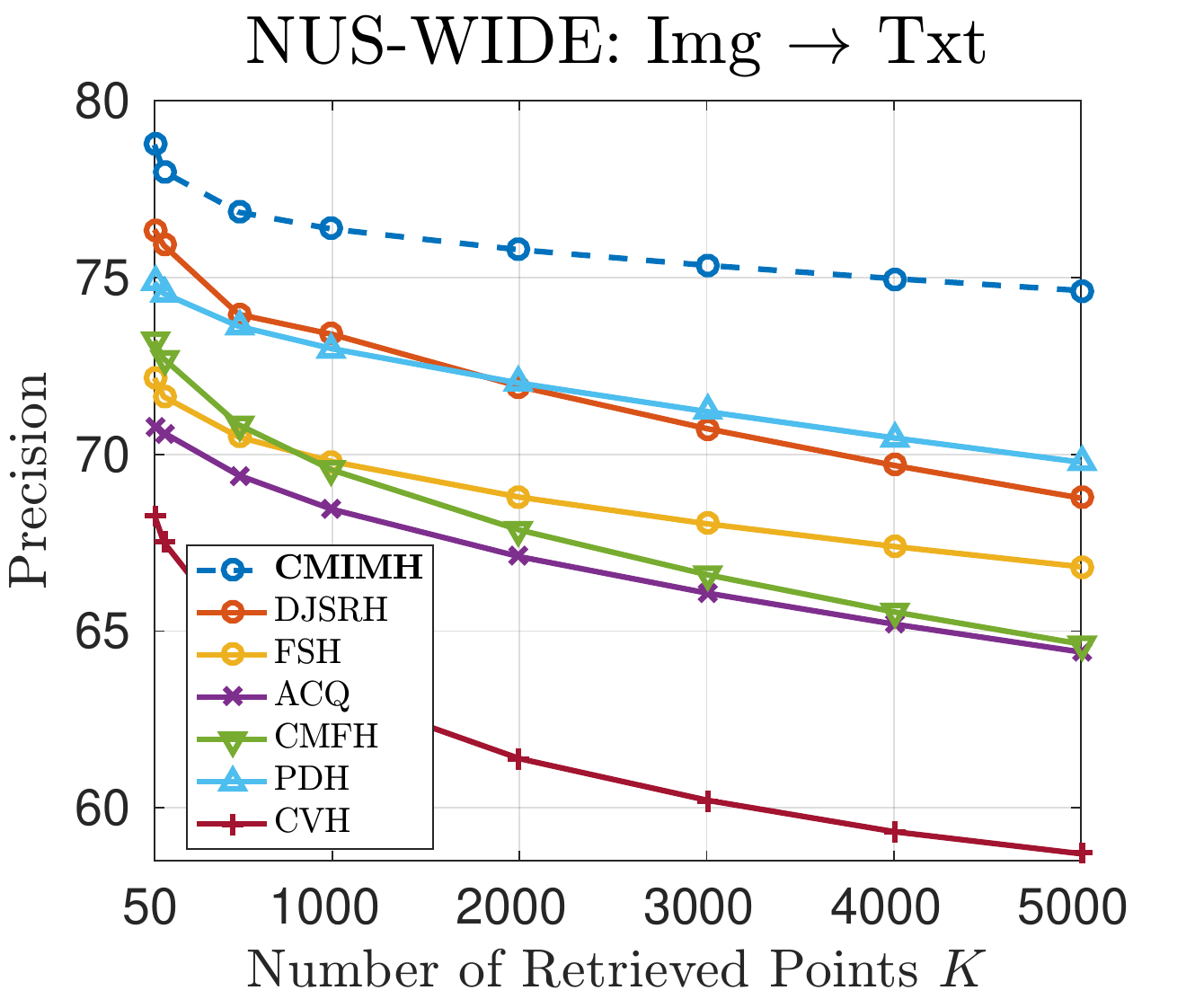}
\includegraphics[width=0.30\textwidth]{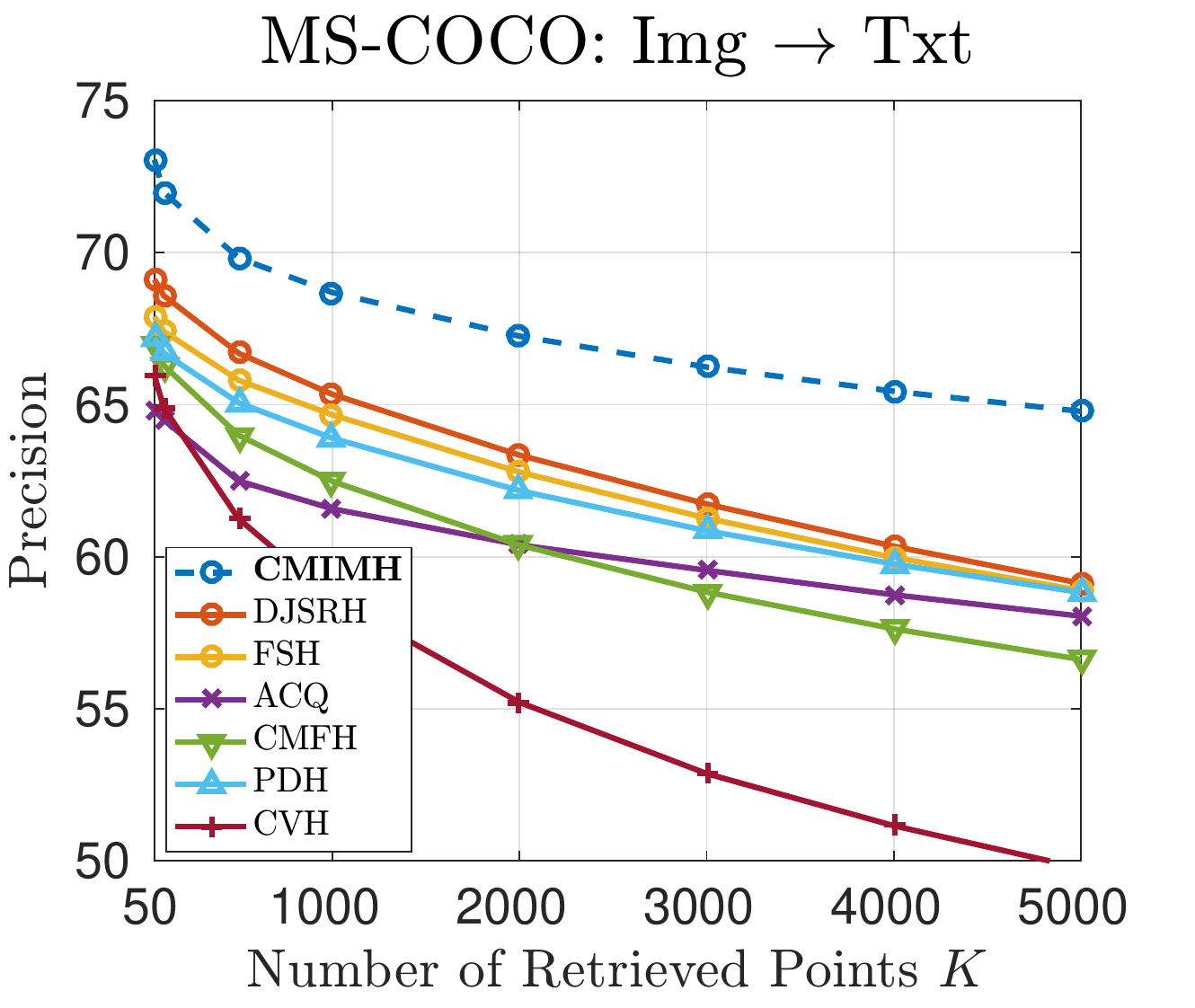}
\end{subfigure}

\vspace{2pt}
\begin{subfigure}[b]{\textwidth}
\centering
\includegraphics[width=0.30\textwidth]{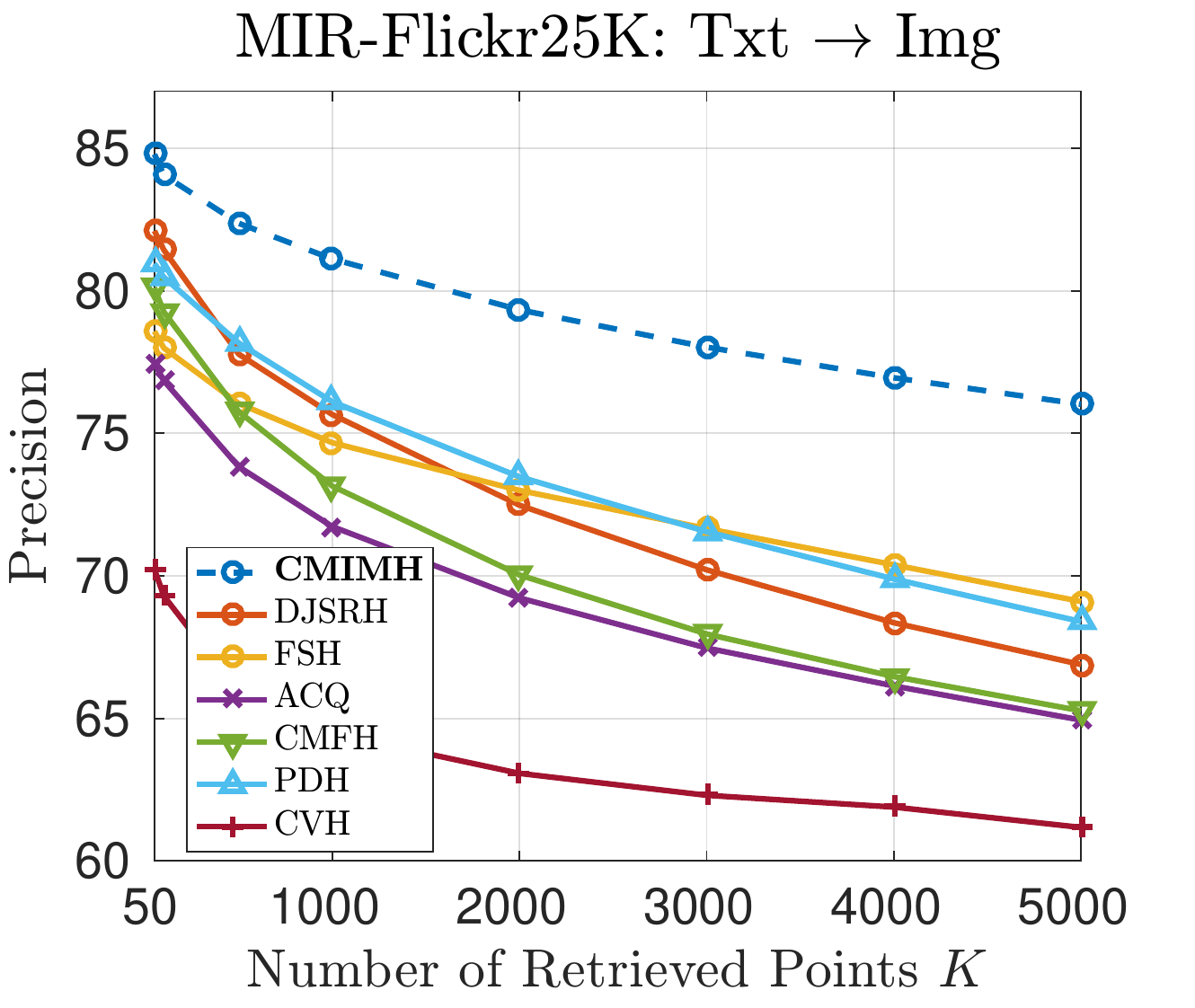}
\includegraphics[width=0.30\textwidth]{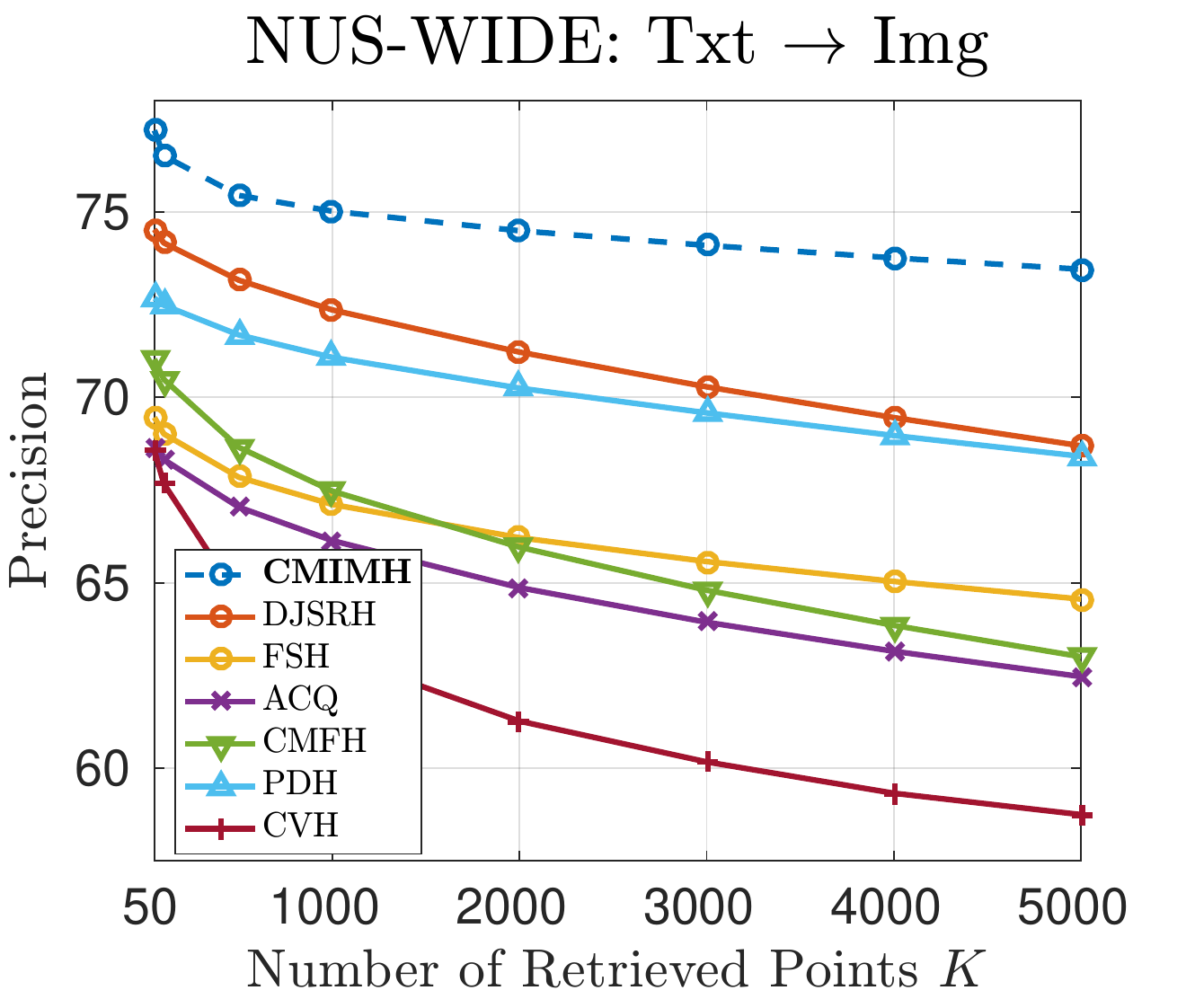}
\includegraphics[width=0.30\textwidth]{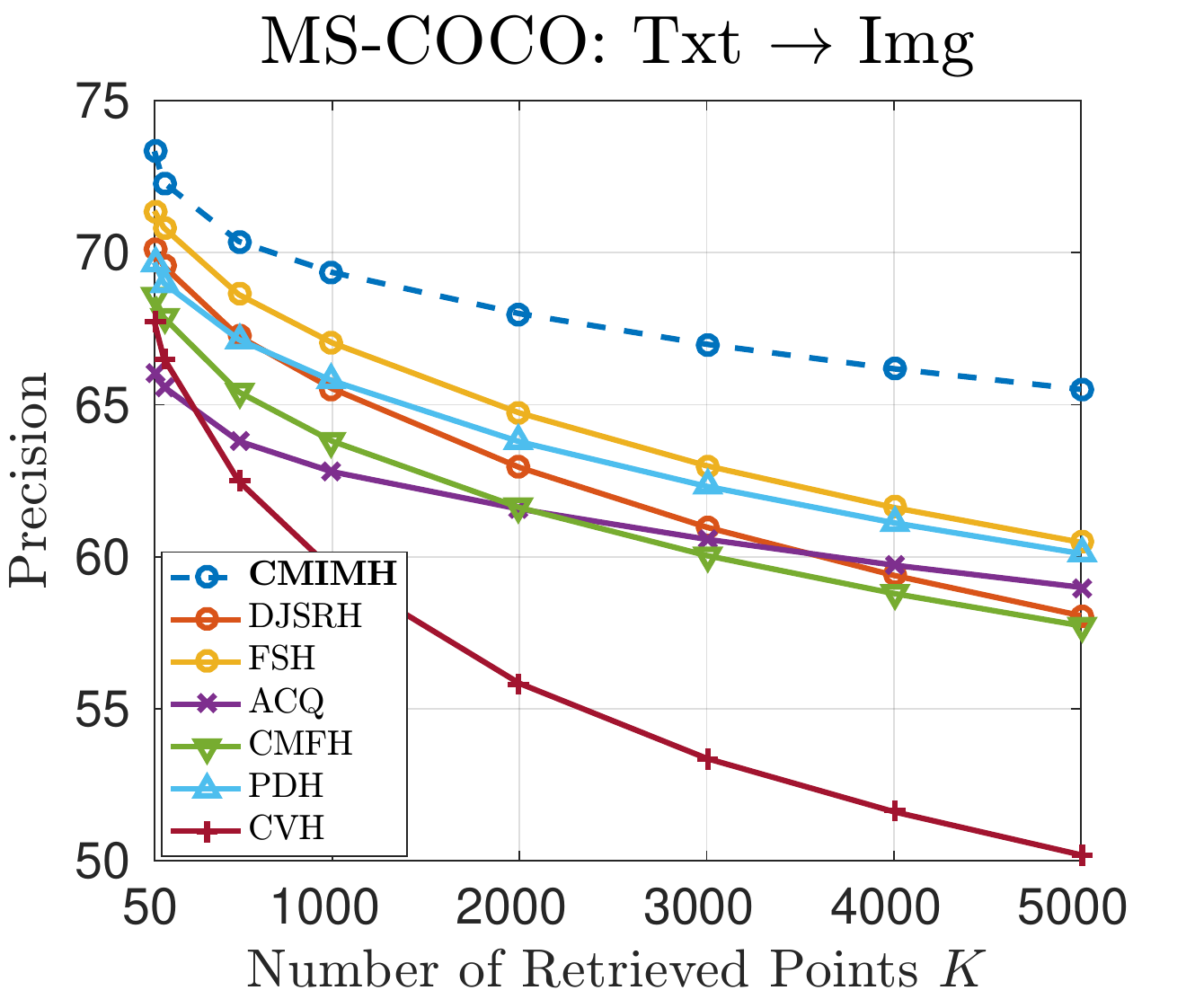}
\end{subfigure}
\vspace{-15pt}
\end{figure*}

\begin{figure*}[t]
\centering
\caption{The Precision-Recall (PR) (\%) curves on different datasets of 32-bit hash codes.}
\label{fig:PRcurve}
\vspace{2pt}
\begin{subfigure}[b]{\textwidth}
\centering
\includegraphics[width=0.30\textwidth]{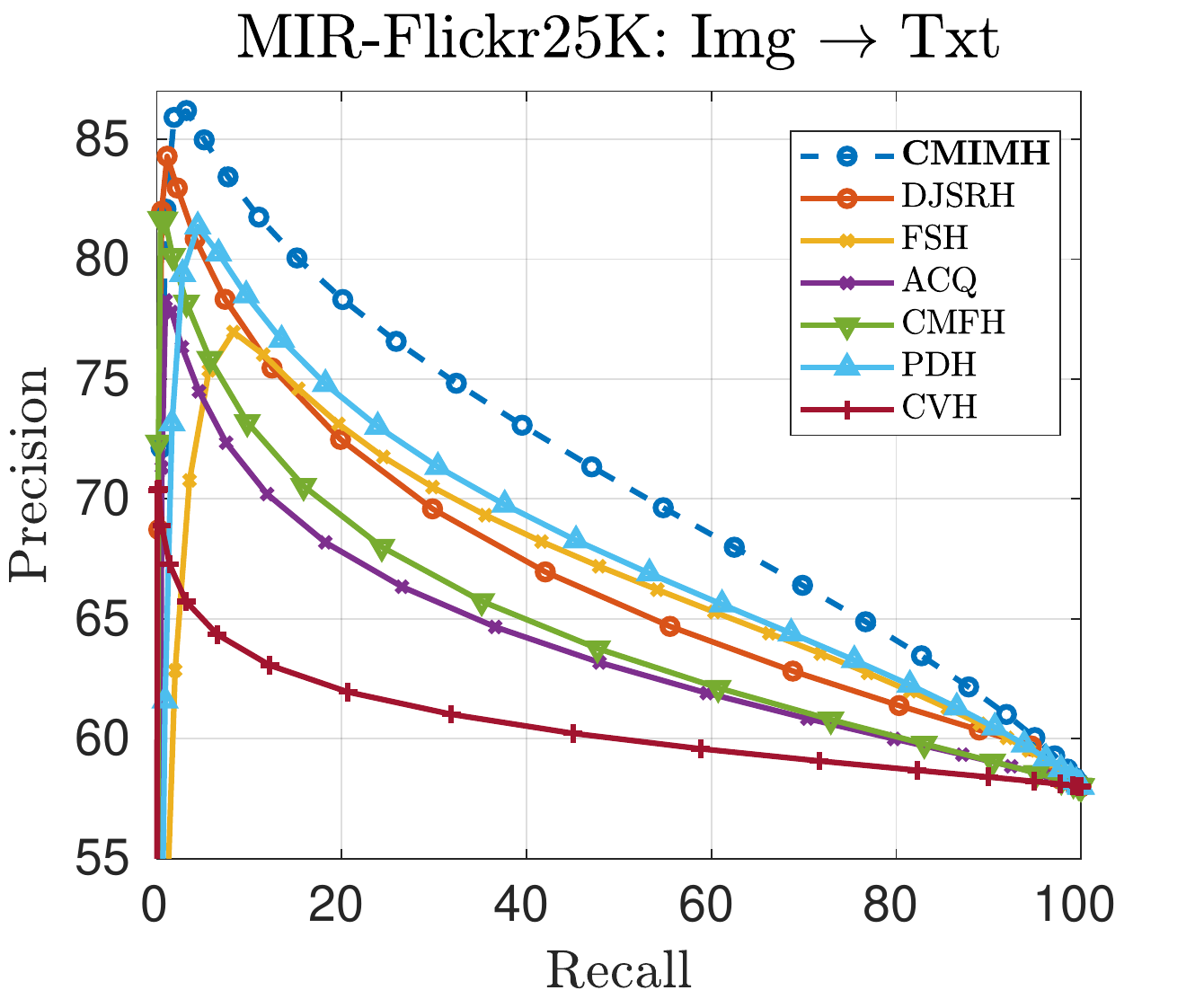}
\includegraphics[width=0.30\textwidth]{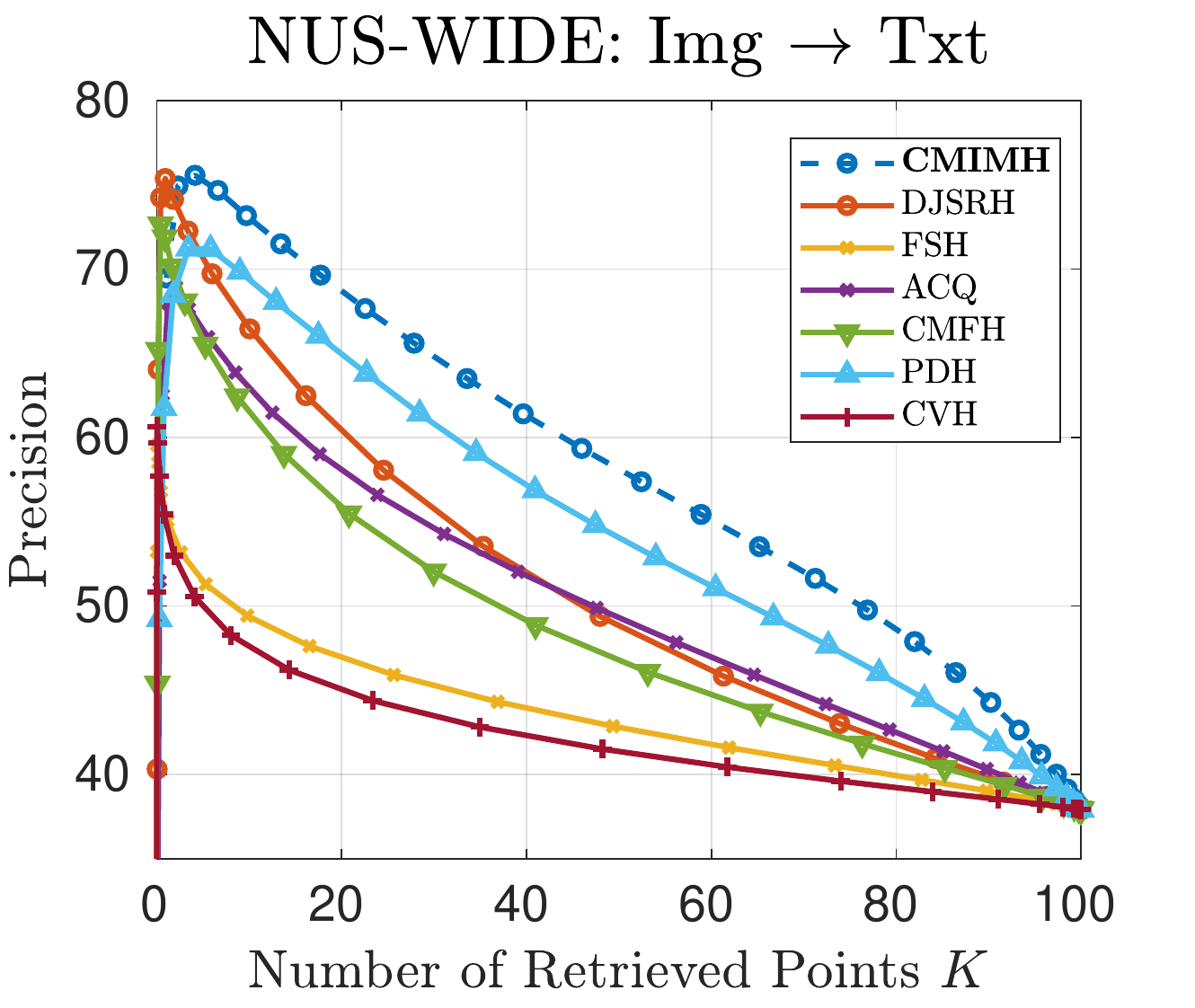}
\includegraphics[width=0.30\textwidth]{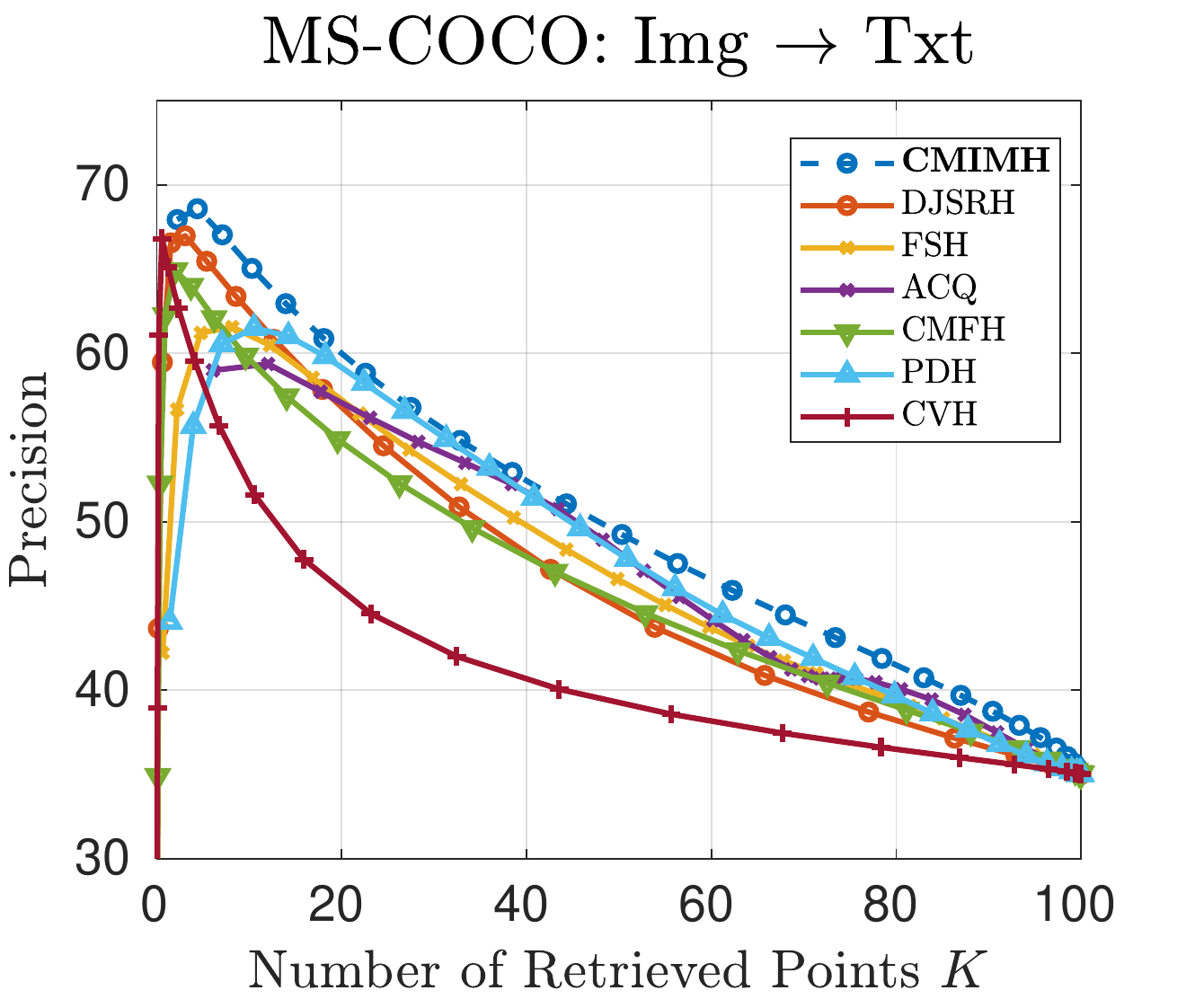}
\end{subfigure}

\vspace{2pt}
\begin{subfigure}[b]{\textwidth}
\centering
\includegraphics[width=0.30\textwidth]{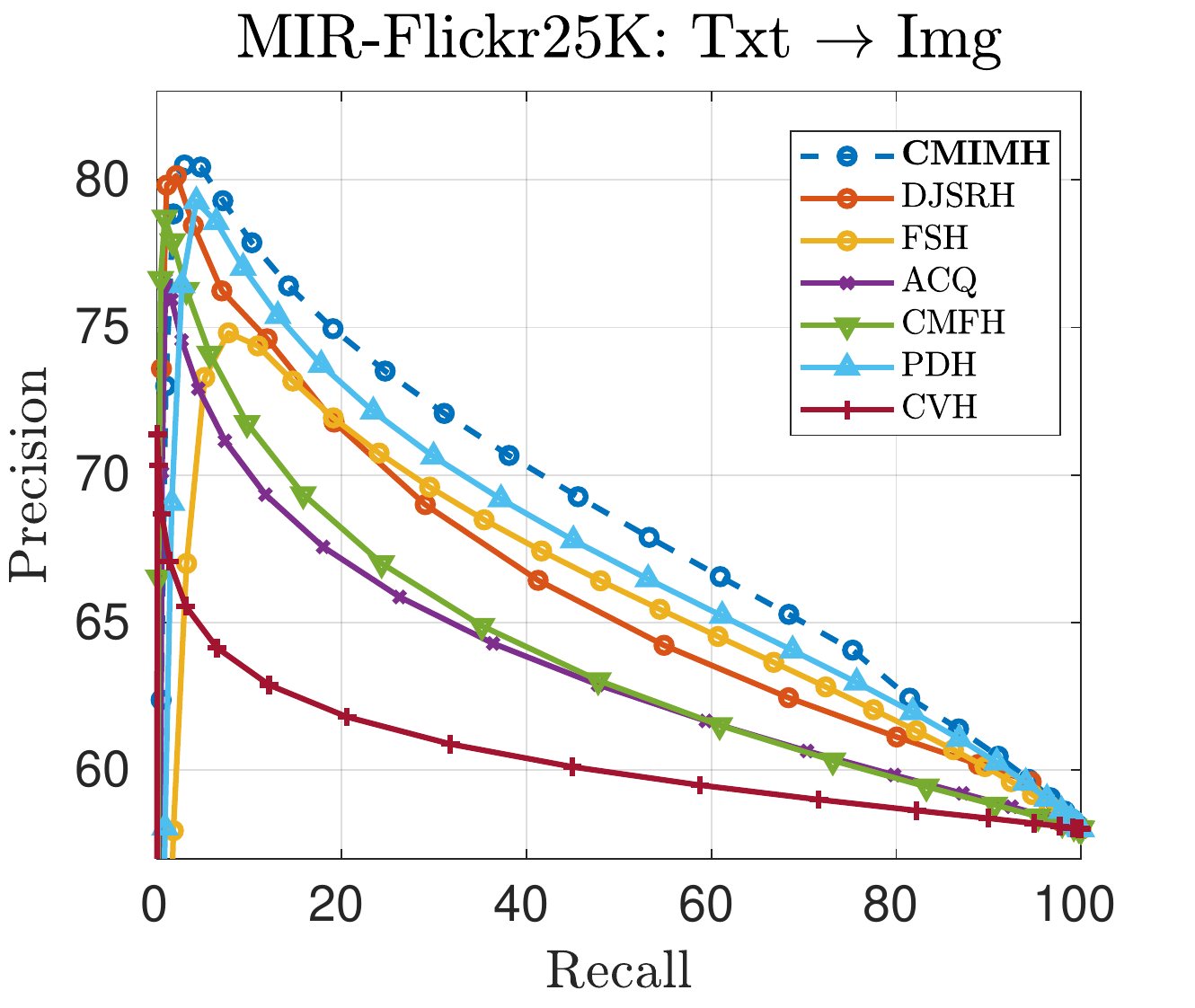}
\includegraphics[width=0.30\textwidth]{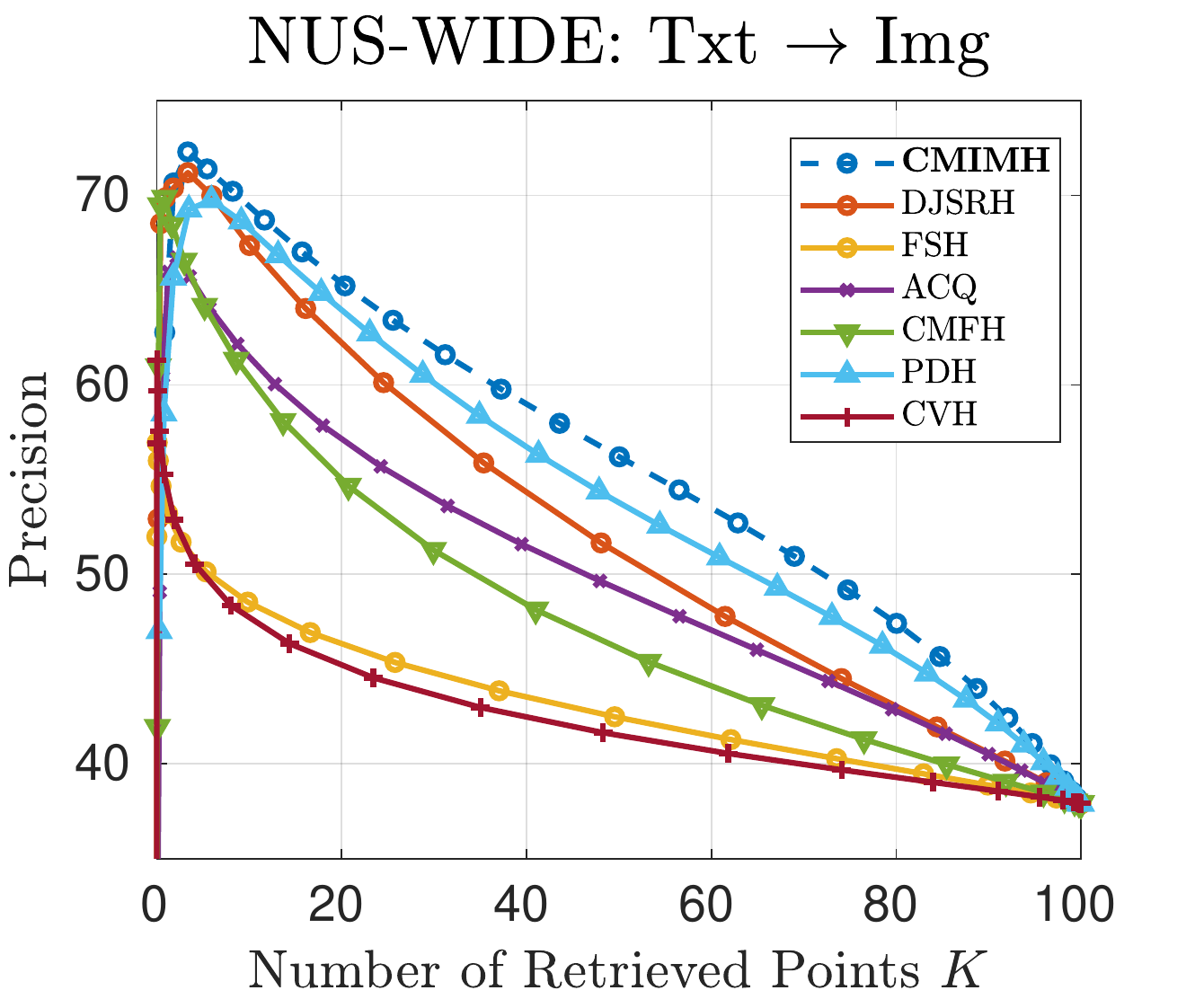}
\includegraphics[width=0.30\textwidth]{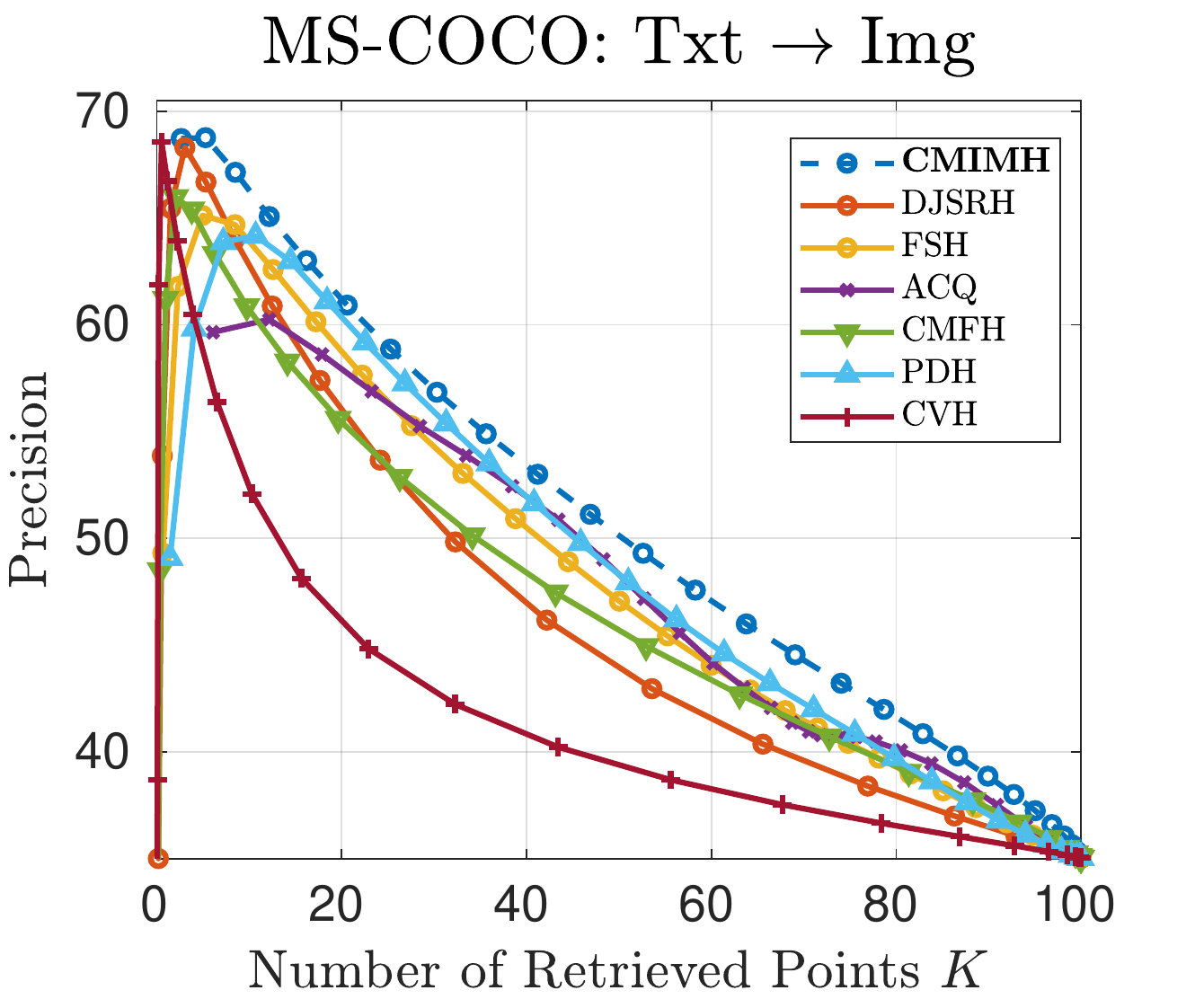}
\end{subfigure}
\vspace{-15pt}
\end{figure*}

\vspace{-4pt}
\subsection{Comparison with the states of the art}
In this section, we compare our proposed method against recent state-of-the-art cross-modal hashing methods, i.e., Cross-View Hashing (CVH) \cite{CVH}, Predictable Dualview Hashing (PDH) \cite{PDH}, Collective Matrix Factorization Hashing (CMFH) \cite{CMFH}, Alternating Co-Quantization (ACQ) \cite{ACQ}, 
Fusion Similarity Hashing (FSH) \cite{FSH},
and Deep Joint Semantics Reconstructing Hashing (DJSRH) \cite{DJSRH}.
From the experimental results in term of \textit{mAP@1k} shown in Table \ref{tb:img_txt}, we can observe that our proposed method outperforms the state-of-the-art unsupervised cross-modal hashing methods including the deep-based method (i.e., DJSRH) at majority of encoding lengths, datasets, and retrieval tasks.
% For the MS-COCO dataset, we achieve comparable performance with DJSRH on the Txt$\to$Txt and Txt$\to$Img retrieval tasks at $L=48$, while still outperforms DJSRH  by clear margins in other settings. We also note that DJSRH requires to fine-tune the very-deep CNN (e.g., AlexNet) for image modality. Hence, it requires much higher computational cost for training.
For the MS-COCO dataset with at $L=48$, CMIMH achieves lower performance than FSH and DJSRH on the Txt$\to$Txt retrieval tasks, while still outperforms DJSRH by clear margins in other settings.

{
Additionally, Figure \ref{fig:pre@K} and Figure \ref{fig:PRcurve}, respectively, show the \textit{Pre@K} curves and Precision-Recall (PR) curves for Img$\to$Txt and  Txt$\to$Img tasks with 32-bit hash codes. Compared with other methods, ours still significantly outperforms the state-of-the-art baselines over the three benchmark datasets for both metrics (i.e., \textit{Pre@K} curves and PR curve). These results confirm the advantages of our proposed method in unsupervised cross-modal retrieval. 
}

\textbf{Comparison with Unsupervised Deep Cross Modal Hashing (UDCMH) \cite{UDCMH}}
Following UDCMH, we report the mAP of top-50 retrieved results (\textit{mAP@50}). The experiment results are shown in Table \ref{tb:compare_UDCMH}. 
We observe that our proposed method can outperform UDCMH by large margins for both MIR-Flickr25k and NUS-WIDE datasets. 

\begin{table}[t]
\small
\centering
\def\arraystretch{1.1}
\caption{{Comparison with UDCMH \cite{UDCMH} using \textit{mAP@50} on MIR-FLickr25k and NUS-WIDE datasets. The results of UDCMH are cited from \cite{UDCMH}.}}
\label{tb:compare_UDCMH}
\resizebox{1.\columnwidth}{!}
{
\begin{tabular}{|c|l|c|c|c|c|c|c|}
\hline
\multirow{2}{*}{Task} & \multirow{2}{*}{Method} & \multicolumn{3}{c|}{MIR-Flickr25k} & \multicolumn{3}{c|}{NUS-WIDE} \\ \cline{3-8}
&  & 16 & 32 & 64 & 16 & 32 & 64 \\ \hline
Img$\to$ & UDCMH  & 68.9 & 69.8 & 71.4 &  51.1 &  51.9 &  52.4  \\ \cline{2-8}
Txt & \textbf{CMIMH} & \textbf{83.2} & \textbf{86.5} & \textbf{88.2} & \textbf{74.52} & \textbf{78.23} & \textbf{79.75} \\ \hline
Txt$\to$ & UDCMH    & 69.2 & 70.4 & 71.8 &  63.7 &  65.3 &  69.5 \\ \cline{2-8}
Img & \textbf{CMIMH} & \textbf{82.4} & \textbf{83.8} & \textbf{86.2} & \textbf{73.89} & \textbf{76.08} & \textbf{79.14} \\ \hline
\end{tabular}
}
\end{table}

\begin{table}[t]
\centering
\def\arraystretch{1.15}
\caption{Comparison with MGAH \cite{MGAH} and {UKD \cite{UKD}} using \textit{mAP} (\%) on MIR-Flickr25k and NUS-WIDE datasets. The results of MGAH and UKD are cited from original papers \cite{MGAH,UKD}.}
\label{tb:compare_ugach}
\resizebox{1.0\columnwidth}{!}{
\begin{tabular}{|c|l|c|c|c|c|c|c|}
\hline
\multirow{2}{*}{Task} & \multirow{2}{*}{Method} & \multicolumn{3}{c|}{MIR-Flickr25k} & \multicolumn{3}{c|}{NUS-WIDE} \\ \cline{3-8}
&  & 16 & 32 & 64 & 16 & 32 & 64 \\ \hline
Img$\to$ & MGAH & 68.5 & 69.3 & 70.4 & {61.3} & 62.3 & 62.8 \\ \cline{2-8}
Txt & UKD-SS & 71.4 & 71.8 & 72.5 & 61.4 & 63.7 & 63.8 \\ \cline{2-8}
 & \textbf{CMIMH} & \textbf{74.15} & \textbf{74.65} & \textbf{75.98} & \textbf{62.72} & \textbf{63.83} & \textbf{64.75} \\ \hline
% Txt & \textbf{CMIMH} & \textbf{00.00} & \textbf{00.00} & \textbf{00.00} & \textbf{00.00} & \textbf{00.00} & \textbf{00.00} \\ \hline
Txt$\to$ & MGAH  & 67.3 & 67.6 & 68.6 & 60.3 & 61.4 & {64.0} \\ \cline{2-8}
Img & UKD-SS & 71.5 & 71.6 & 72.1 & \textbf{63.0} & \textbf{65.6} & \textbf{65.7} \\ \cline{2-8}
 & \textbf{CMIMH} & \textbf{73.87} & \textbf{74.24} & \textbf{75.41} & {62.41} & {63.70} & {64.23}\\ \hline
% Img & \textbf{CMIMH} & \textbf{00.00} & \textbf{00.00} & \textbf{00.00} & \textbf{00.00} & \textbf{00.00} & \textbf{00.00} \\ \hline
\end{tabular}
% \vspace{-1.2em}
}
\end{table}

% \smallskip
{
\textbf{Comparison with Multi-pathway Generative Adversarial Hashing (MGAH) \cite{MGAH} and Unsupervised Knowledge Distillation for Cross-Modal Hashing (UKD) \cite{UKD}:}
We conduct additional experiments on MIR-FLickr25k and NUS-WIDE datasets. For a fair comparison, we follow the experiment settings from \cite{MGAH} and \cite{UKD}. Specifically, the FC7 features of the pretrained 19-layer VGGNet are used for images. 1,000-dimension BOW features are used for texts in both datasets. 1\% samples of the NUS-WIDE dataset and 5\% samples of the MIR-FLickr25k dataset are used as the query
sets, and the rest as training set and also the retrieval database. 
% For the very large dataset NUS-WIDE, we randomly select 20,000 pairs from the database to form the training set for our method.
% In addition, to avoid biases due to random-sample query sets, we repeat our methods 10 times with different query sets and report the average values. 
We present the retrieval performance in term of \textit{mAP} (of all retrieved samples) are shown in Table \ref{tb:compare_ugach}.
We can observe that our proposed CMIMH consistently outperforms MGAH for both MIR-Flickr25k and NUS-WIDE datasets, especially for MIR-Flickr25k where the improvement gaps are greater than $5\%$. 
In comparison with UKD \cite{UKD}, our proposed method achieves lower performance on NUS-WIDE dataset - Txt$\to$Img task, while still outperform UKD on NUS-WIDE dataset - Img$\to$Txt task and MIR-Flickr25k for both tasks. These results show that our proposed method is still more favorable than UKD.
% For NUS-WIDE dataset, even though using less training samples, our proposed DCSH generally outperforms UGACH for the majority of cases.
}

{
\textbf{Comparison with Learning Disentangled Representation for Cross-Modal Retrieval with Deep Mutual Information Estimation (LDR) \cite{disentangle_rep_with_MI}:}
To have a fair comparison, we follow the experiment setting of \cite{disentangle_rep_with_MI}. Specifically, we extract VGG19 FC7 feature for images. We randomly select 1000 images as query, 1000 images as validation and the remaining images as database as well as training set.
We report in Table \ref{tb:compare_LDR} the retrieval performance in term of \textit{RecallOne@K}, the percent of queries for which the ground-truth is one of the first $K$ retrieved.
% Additionally, to avoid bias due to the random test sets, we conduct experiments 10 times with 10 different test sets and report the average $\pm$ standard deviation in Table \ref{tb:compare_LDR}.
In \cite{disentangle_rep_with_MI}, the authors reported the \textit{RecallOne@K} for 1024-dimension real-value features (32,768 bits), we find out that even with 128 bit hash codes, our proposed method can outperform LDR by large margins for both \textit{RecallOne@1} and \textit{RecallOne@10}.

\begin{table}[h]
\centering
\def\arraystretch{1.1}
\caption{{Comparison with LDR \cite{disentangle_rep_with_MI} using \textit{RecallOne@K} (\%) on MS-COCO. The results of LDR are cited from \cite{disentangle_rep_with_MI}.}}
\label{tb:compare_LDR}
% \resizebox{1.0\columnwidth}{!}
{
\begin{tabular}{|l|c|c|c|c|c|}
\hline
\multirow{2}{*}{Method} & \multicolumn{2}{c|}{Img$\to$Txt} & \multicolumn{2}{c|}{Img$\to$Txt} \\ \cline{2-5}
& $K=1$ & $K=10$ & $K=1$ & $K=10$ \\ \hline
% LDR \cite{disentangle_rep_with_MI} & \multirow{2}{*}{53.4} & \multirow{2}{*}{91.3} & \multirow{2}{*}{40.5} & \multirow{2}{*}{88.7} \\   \hline
% (1024-D) & & & & \\ \hline
LDR (1024-D) & {53.4} & {91.3} & {40.5} & {88.7} \\   \hline
%  & & & & \\ \hline
\textbf{CMIMH} (128 bits) & \textbf{65.53}  & \textbf{94.35} & \textbf{69.82}  & \textbf{93.32}  \\  \hline
%  & ($\pm$1.34) &  ($\pm$0.52) & ($\pm$1.46) & ($\pm$0.66) \\ \hline

\end{tabular}
% \vspace{-1.2em}
}
\end{table}
}

\paragraph{Comparison with state of the art using hand-crafted image feature}
Following the experiment setting of CRE \cite{CRE} and FSH \cite{FSH}, we conduct experiments with hand-crafted features on NUS-WIDE datasets. For NUS-WIDE dataset, each image is represented by 500-dimensional BoW SIFT features and each text is represented by a 1,000-dimension preprocessed BOW feature. We randomly select 2,000 pairs as the query set; the remaining are used as the database. We also sample 20,000 pairs from the database as the training set. We present the experiment results in terms of \textit{mAP} (of all returned samples) and \textit{Pre@100} (as used in \cite{FSH,CRE}) in Table \ref{tb:compare_hand_crafted}. 
{We can observe that our proposed method can also work well with hand-crafted features and outperforms all compared methods.}
\begin{table}[t]
% \small
\centering
\def\arraystretch{1.1}
\setlength{\tabcolsep}{7pt}
\caption{Comparison using \textit{mAP} (\%) (of all returned samples) and \textit{Pre@100} on NUS-WIDE using hand-crafted features for images.
The results of CRE method are cited from \cite{CRE} and the results of other methods are cited from \cite{FSH}.
}
\label{tb:compare_hand_crafted}
\resizebox{1.0\columnwidth}{!}
{
\begin{tabular}{|c|c|c|c|c|c|c|}
\hline
\multirow{3}{*}{Task} & \multirow{3}{*}{Method} & \multicolumn{5}{c|}{NUS-WIDE} \\ \cline{3-7}
& & \multicolumn{3}{c|}{\textit{mAP}} & \multicolumn{2}{c|}{\textit{Pre@100}} \\ \cline{3-7}
&  & 16 & 32 & 64 & 32 & 64  \\ \hline
\multirow{6}{*}{\rotatebox{90}{Img$\to$Txt}} 
& CVH  & 38.11 & 36.85 & 35.74 & 47.49 & 43.87 \\  \cline{2-7}
& PDH  & 46.58 & 47.47 & 47.80 & 49.89 & 51.25 \\ \cline{2-7}
& CMFH & 37.23 & 37.81 & 37.99 & 50.64 & 53.09 \\ \cline{2-7}
& ACQ  & 42.47 & 44.35 & 43.28 & 44.42 & 43.09 \\ \cline{2-7}
& FSH \cite{FSH}  & 50.59 & 50.63 & 51.71 & 52.97 & 56.16 \\ \cline{2-7}
& CRE \cite{CRE}  & 51.31 & 52.99 & 53.32 & - & - \\ \cline{2-7}
& \textbf{CMIMH} & \textbf{53.66} & \textbf{53.95} & \textbf{55.34} & \textbf{63.98} & \textbf{65.68}  \\ \hline
\multirow{6}{*}{\rotatebox{90}{Txt$\to$Img}} 
& CVH  & 37.68 & 36.52 & 35.55 & 46.90 & 43.21 \\ \cline{2-7}
& PDH  & 44.58 & 45.19 & 45.52 & 51.33 & 52.84 \\ \cline{2-7}
& CMFH & 39.57 & 40.36 & 41.05 & 45.17 & 45.95 \\ \cline{2-7}
& ACQ  & 41.34 & 42.73 & 42.00 & 45.73 & 48.87 \\ \cline{2-7}
& FSH  & 47.90 & 48.10 & 49.65 & 53.88 & 56.85 \\ \cline{2-7}
& CRE  & 49.27 & 50.86 & 51.49 & - & - \\ \cline{2-7}
& \textbf{CMIMH}   & \textbf{53.01} & \textbf{53.44} & \textbf{55.19} & \textbf{60.77} & \textbf{61.37} \\ 
\hline
\end{tabular}
}
\end{table}

\paragraph{Effect of training size}
Different from most of the state-of-the-art methods \cite{CVH,PDH,CMFH,ACQ,FSH,UDCMH}, our proposed method can be fully-optimized using gradient descent in a mini-batch manner. Therefore, our proposed method can be easily trained with much larger training sets.
We further analyze the effects on retrieval performance when varying the training size on NUS-WIDE dataset. We also compare our retrieval performance with the retrieval performance of DJSRH \cite{DJSRH}, which is one of our most competitive methods and also can be trained in a mini-batch manner.
% to emphasize the advantage of our proposed method. 
The retrieval performance in term of \textit{mAP@1k} when using different training set sizes are shown in Table \ref{tb:training_size}. We can observe that our proposed method can achieve higher \textit{mAP@1k} when utilizing more training data. Furthermore, our proposed method also consistently outperforms DJSRH \cite{DJSRH} with different training sizes. %We note that DJSRH needs to construct a similarity matrix for each mini-batch
\begin{table}[t]
\small
\centering
\def\arraystretch{1.1}
\setlength{\tabcolsep}{6pt}
\caption{The cross-modal retrieval performance in term of \textit{mAP@1k} (\%) on the NUS-WIDE dataset with different numbers of training samples. 
}
\label{tb:training_size}
\resizebox{1.0\columnwidth}{!}
{
\begin{tabular}{|c|c|c|c|c|c|}
\hline
\multirow{2}{*}{L} & \multirow{2}{*}{Task} & \multirow{2}{*}{Method} & \multicolumn{3}{c|}{Training size} \\ \cline{4-6}
& & & 5k & 10k & 20k \\ \hline

% \hline\hline
\multirow{4}{*}{16} & \multirow{2}{*}{ Img$\to$Txt} 
 & DJSRH \cite{DJSRH} & 71.23 & 72.67 & 73.46 \\  \cline{3-6}
& & \textbf{CMIMH} & \textbf{73.59} & \textbf{75.14} & \textbf{77.25} \\  \cline{2-6}
 & \multirow{2}{*}{ Txt$\to$Img} 
 & DJSRH & 68.18 & 71.12 & 73.53 \\  \cline{3-6}
& & \textbf{CMIMH} & \textbf{73.39} & \textbf{75.45} & \textbf{76.54} \\  \hline

% \hline\hline
\multirow{4}{*}{32} & \multirow{2}{*}{ Img$\to$Txt} 
 & DJSRH & 74.86 & 76.93 & 77.58 \\  \cline{3-6}
& & \textbf{CMIMH} & \textbf{76.20} & \textbf{77.80} & \textbf{79.06} \\  \cline{2-6}
 & \multirow{2}{*}{ Txt$\to$Img} 
 & DJSRH & 73.29 & 74.97 & 77.31 \\  \cline{3-6}
& & \textbf{CMIMH} & \textbf{74.88} & \textbf{76.46} & \textbf{78.23} \\  \hline

% \hline\hline
\multirow{4}{*}{48} & \multirow{2}{*}{ Img$\to$Txt} 
 & DJSRH & 76.52 & 78.17 & 78.78 \\  \cline{3-6}
& & \textbf{CMIMH} & \textbf{76.71} & \textbf{78.62} & \textbf{79.66} \\  \cline{2-6}
 & \multirow{2}{*}{ Txt$\to$Img} 
 & DJSRH & 74.72 & 77.35 & 77.29 \\  \cline{3-6}
& & \textbf{CMIMH} & \textbf{75.98} & \textbf{78.12} & \textbf{78.81} \\  \hline
\end{tabular}
}
\vspace{-0.5em}
\end{table}

\vspace{-1pt}
\section{Conclusion}
In this paper, inspired by recent advances in learning representation by maximizing mutual information, we proposed a novel framework, dubbed Cross-Modal Info-Max Hashing (CMIMH). By assuming the binary representations to be modeled by multivariate Bernoulli distributions, we can maximize the MI effectively using gradient descent optimization in a mini-batch manner via maximizing their estimated variational lower-bounds. We additionally find out that trying to minimize modality gap by learning similar binary representations for the same instance from different modalities could result in  modality-private information loss. Properly balancing the modality gap and modality-private information loss is important to achieve better performance.
Experiment results confirm the effectiveness of our proposed method for both cross-modal and single-modal retrieval tasks.
Additionally, the ablation studies clearly justify the advantages of different components in our proposed method.

% Finally, in this paper, following the common setting in recent state-of-the-art methods \cite{UDCMH,DJSRH} for cross-modal retrieval, we only discussed and evaluated our proposed method for two modalities, e.g., image and text. However, we believe that our proposed method can be adopted for the general cases of $M$-modalities ($M\ge 2$). We leave this extension 
% % for the general cases of $M$-modalities 
% for future works.

%\appendices

%% use section* for acknowledgment
%\ifCLASSOPTIONcompsoc
%  % The Computer Society usually uses the plural form
%\else
%  % regular IEEE prefers the singular form
%\fi

% \ifCLASSOPTIONcaptionsoff
%   \newpage
% \fi

\section*{Acknowledgement}

This project was supported by SUTD project PIE-SGP-AI-2018-01. This research was also supported by the National Research Foundation Singapore under its AI Singapore Programme [Award Number:AISG-100E2018-005].

%\vspace{-0.6em}
\begin{table*}[t]
\begin{appendices}
%\clearpage
% \onecolumn

\section{More detail derivation of Equation \eqref{eq:private_info_upper_bound}}
\label{appendix:detail_upper_bound}

\begin{equation}
\tag{\ref{eq:private_info_upper_bound}}
\begin{split}
% \label{eq:private_info_upper_bound}
I(\bxi; \bhi|\bxt)&= \bbE_{\bsxi;\bsxt\sim p(\bxi,\bxt)}\bbE_{\bsh\sim P_{\theta_i} (\bhi|\bxi)} \left[\log \frac{p(\bxi = \bsxi,\bhi = \bsh|\bxt = \bsxt)}{p(\bxi = \bsxi|\bxt = \bsxt) P_{\theta_i} (\bhi = \bsh|\bxt = \bsxt)}\right] \\
&= \bbE_{\bsxi;\bsxt\sim p(\bxi,\bxt)}\bbE_{\bsh\sim P_{\theta_i} (\bhi|\bxi)} \left[\log \frac{p(\bxi = \bsxi|\bxt = \bsxt) P_{\theta_i}(\bhi = \bsh|\bxi = \bsxi,\bxt = \bsxt)}{p(\bxi = \bsxi|\bxt = \bsxt) P_{\theta_i} (\bhi = \bsh|\bxt = \bsxt)}\right]  \\
&= \bbE_{\bsxi;\bsxt\sim p(\bxi,\bxt)}\bbE_{\bsh\sim P_{\theta_i} (\bhi|\bxi)} \left[\log \frac{P_{\theta_i} (\bhi = \bsh|\bxi = \bsxi)}{P_{\theta_i} (\bhi = \bsh|\bxt = \bsxt)}\right]  \\
&= \bbE_{\bsxi;\bsxt\sim p(\bxi,\bxt)}\bbE_{\bsh\sim P_{\theta_i} (\bhi|\bxi)} \left[\log \frac{P_{\theta_i} (\bhi = \bsh|\bxi = \bsxi)}{P_{\theta_t} (\bht = \bsh|\bxt = \bsxt)}\right] \\
&\qquad\qquad\qquad\qquad+ \bbE_{\bsxi;\bsxt\sim p(\bxi,\bxt)}\bbE_{\bsh\sim P_{\theta_i} (\bhi|\bxi)} \left[\log \frac{P_{\theta_t} (\bht = \bsh|\bxt = \bsxt)}{P_{\theta_i} (\bhi = \bsh|\bxt = \bsxt)}\right]  \\
&= \DKL\left[P_{\theta_i} (\bhi|\bxi)||P_{\theta_t} (\bht|\bxt)\right] - \DKL\left[P_{\theta_i} (\bht|\bxi)||P_{\theta_t} (\bht|\bxt)\right] \\
&\le \DKL\left[P_{\theta_i} (\bhi|\bxi)||P_{\theta_t} (\bht|\bxt)\right].
\end{split}
\end{equation}
Note that since $\bhi$ is completely determined by $\bxi$, $H(\bhi|\bxi)=0$.
% Note that, since $H(\bhi|\bxi)=0$; 
Consequently,
$\bhi$ and $\bxt$ are conditionally independent given $\bxi$, i.e.,
\begin{equation}
P_{\theta_i}(\bhi = \bsh|\bxi = \bsxi,\bxt = \bsxt)=P_{\theta_i}(\bhi = \bsh|\bxi = \bsxi)   
\vspace{-1em}
\end{equation}

% \end{table*}

% \begin{table}[]
%     \centering
%     \begin{tabular}{c|c}
%          &  \\
%          & 
%     \end{tabular}
%     \caption{Caption}
%     \label{tab:my_label}
% \end{table}

% \begin{equation}
%     \bsxi,\bsxt\sim p(\bxi, \bxt)
% \end{equation}
% \begin{equation}
%     \bshi\sim P(\bhi|\bxi=\bsxi)
% \end{equation}
% \begin{equation}
%     \bsht\sim P(\bht|\bxt=\bsxt)
% \end{equation}

% \begin{equation}
%     p(\bhi,\bht) = p(\bxi, \bxt)P(\bhi|\bxi)P(\bht|\bxt)
% \end{equation}
% \begin{equation}
%     p(\bhi) = p(\bxi)P(\bhi|\bxi)
% \end{equation}
% \begin{equation}
%     p(\bht) = p(\bxt)P(\bht|\bxt)
% \end{equation}
\end{appendices}
\end{table*}
{%\small
\bibliographystyle{IEEEtran}
\bibliography{hash}
}

\end{document}